\documentclass[10pt,twocolumn,letterpaper]{article}

\usepackage[pagenumbers]{cvpr} %

\definecolor{cvprblue}{rgb}{0.21,0.49,0.74}
\usepackage[pagebackref,breaklinks,colorlinks,allcolors=cvprblue]{hyperref}
\usepackage{multirow}
\usepackage{colortbl}
\usepackage{pifont}
\usepackage{booktabs} %
\usepackage{amssymb} %
\usepackage{graphicx} %
\usepackage{arydshln}
\usepackage{pifont}

\def\paperID{18500} %
\def\confName{CVPR}
\def\confYear{2025}

\title{CrossVideoMAE: Self-Supervised Image-Video Representation Learning with Masked Autoencoders}
\vspace{-20pt}

\author{
Shihab Aaqil Ahamed$^{1,2*}$, Malitha Gunawardhana$^{3*}$, Liel David$^{4}$, Michael Sidorov$^{4}$, \\ Daniel Harari$^{4}$, Muhammad Haris Khan$^{2}$\\
$^1$Dept. of Electronic and Telecommunication Engineering, University of Moratuwa, Sri Lanka\\
$^2$Mohamed Bin Zayed University of Artificial Intelligence, UAE\\
$^3$University of Auckland, New Zealand,
$^4$Weizmann Institute of Science, Israel\\
\tt\small{\href{mailto:ahamedmbsa.20@uom.lk}{
ahamedmbsa.20@uom.lk}}
}

\begin{document}
\maketitle

\def \thefootnote{*}\footnotetext{Equally contributing authors}

\begin{abstract} 

Current video-based Masked Autoencoders (MAEs) primarily focus on learning effective spatiotemporal representations from a visual perspective, which may lead the model to prioritize general spatial-temporal patterns but often overlook nuanced semantic attributes like specific interactions or sequences that define actions - such as action-specific features that align more closely with human cognition for space-time correspondence. This can limit the model's ability to capture the essence of certain actions that are contextually rich and continuous.
Humans are capable of mapping visual concepts, object view invariance, and semantic attributes available in static instances to comprehend natural dynamic scenes or videos. Existing MAEs for videos and static images rely on separate datasets for videos and images, which may lack the rich semantic attributes necessary for fully understanding the learned concepts, especially when compared to using video and corresponding sampled frame images together. To this end, we propose CrossVideoMAE an end-to-end self-supervised cross-modal contrastive learning MAE that effectively learns both video-level and frame-level rich spatiotemporal representations and semantic attributes. Our method integrates mutual spatiotemporal information from videos with spatial information from sampled frames within a feature-invariant space, while encouraging invariance to augmentations within the video domain. This objective is achieved through jointly embedding features of visible tokens and combining feature correspondence within and across modalities, which is critical for acquiring rich, label-free guiding signals from both video and frame image modalities in a self-supervised manner. Extensive experiments demonstrate that our approach surpasses previous state-of-the-art methods and ablation studies validate the effectiveness of our approach. 

\end{abstract}

\vspace{-8pt}
\section{Introduction}
\label{sec:intro}

\begin{figure}[t!]
\centering
\includegraphics[width=\linewidth]{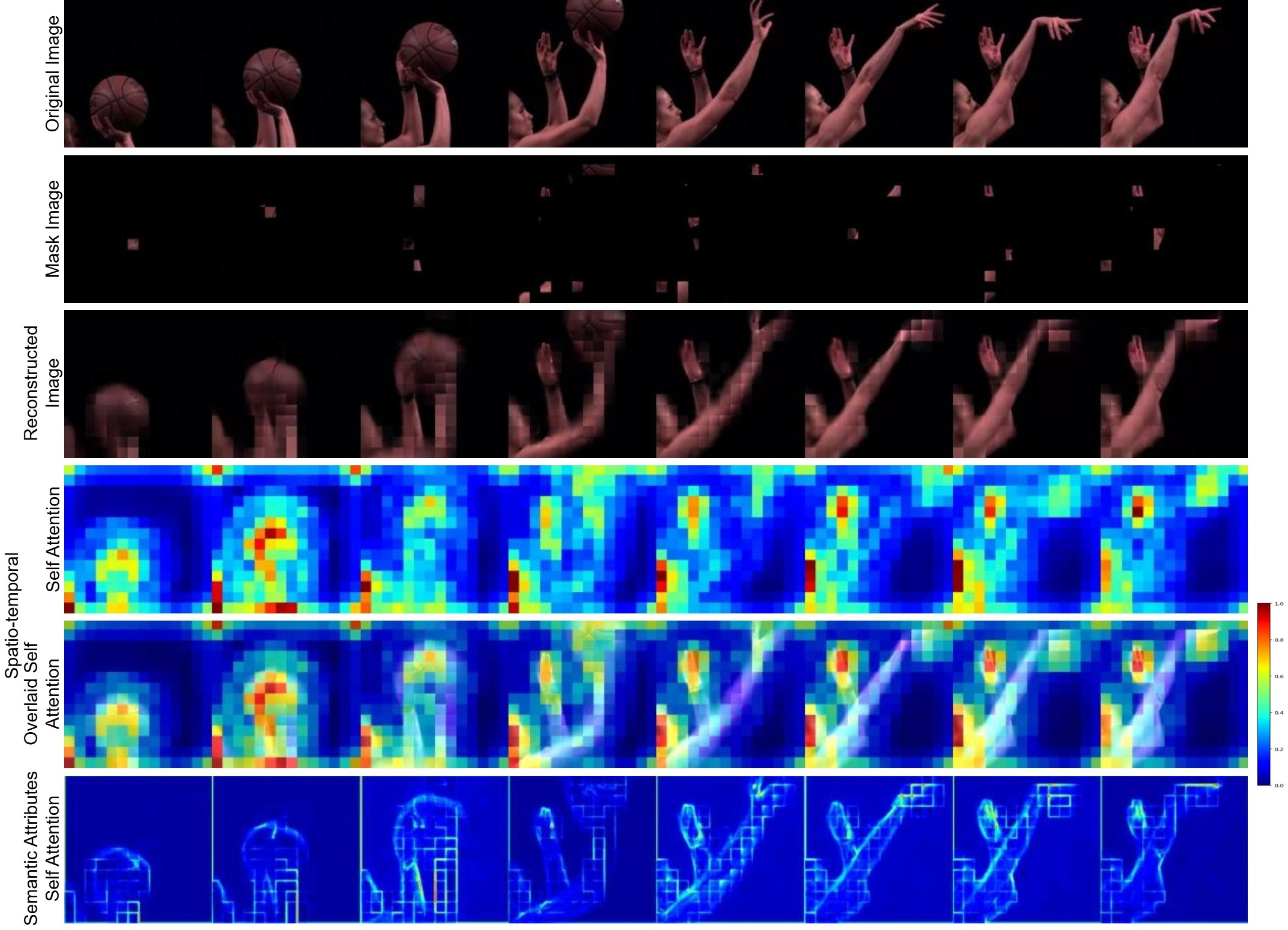}
\captionsetup{font=footnotesize}
\caption{\textbf{Self-attention maps visualization of the proposed approach.}. This demonstrates the efficacy of our method in learning spatiotemporal and semantic representations. The rows depict: original video frames from an action video sequence (\textit{first row}), masked frames with random masking applied (\textit{second row}), reconstructed frames (\textit{third row}), self-attention heatmaps highlighting spatiotemporal representations (\textit{fourth row}), overlaid self-attention heatmaps on reconstructed frames (\textit{fifth row}), and semantic self-attention maps visualizing semantic attributes(\textit{sixth row}). Our approach aim to capture spatiotemporal-spatial feature embedding correspondence of visible tokens across sampled frames and videos,  utilizing differences
between masking ratios (90\% and 95\%), to relate high-level visual and semantic tokens that encode intricate relationships. This joint intra-modal and cross-modal feature embedding at both video and frame level settings enhances invariance to augmentations in the video domain and facilitates effective semantic knowledge distillation from sampled frames to videos. \textit{(ref
supplementary for more visualizations.)}}
\label{fig:report}
\vspace{-20pt}
\end{figure}

Self-supervised learning (SSL) has become a game-changer in reducing reliance on labeled data by harnessing vast quantities of unlabeled information to derive meaningful representations. This approach has shown great promise across domains, including images~\cite{caron2021emerging, radford2021learning, morgado2021audio} and videos~\cite{Jwang2020self, gunawardhana2024effective, dave2024unifying}, establishing a strong foundation for various downstream applications. In particular, video-based action recognition has emerged as an important field with implications for intelligent surveillance, healthcare, and human-computer interaction~\cite{tran2015learning, wang2021tdn, wang2018temporal,zhang2021co}. However, action recognition in video continues to face challenges unique to this medium, such as substantial data redundancy, scene diversity, and the high cost of labeling extensive video datasets~\cite{kumar2023large, kim2024fusion}.

Recent SSL advances for video understanding have partially addressed these challenges through generative models~\cite{chen2020generative, esser2021taming}, reconstruction~\cite{vondrick2016generating,yao2020video} and specialized pretext tasks~\cite{jing2018self, yao2020video, wang2021unsupervised}. Furthermore, contrastive learning, proven highly effective in image-based SSL~\cite{kalantidis2020hard, grill2020bootstrap, he2020momentum, misra2020self}, has been adapted to video data~\cite{huang2021self, qian2021spatiotemporal, pan2021videomoco}. However, many video-based contrastive methods focus on augmentation invariance while overlooking the potential of cross-modal learning, which has demonstrated promise in enriching SSL by integrating complementary data streams from multiple modalities, such as text-image~\cite{desai2021virtex, radford2021learning, sariyildiz2020learning} and audio-video~\cite{morgado2021robust, korbar2018cooperative}.

Existing image- and video-based Masked Autoencoders (MAEs)~\cite{wang2022bevt,girdhar2023omnimae} primarily align low-level visual features with semantic attributes, yet often overlook the intrinsic correlation between sequential frames in video. For example, an existing video-based MAE pre-trained model may misinterpret a Fig.~\ref{fig:report} basketball shot action as a mere arm movement, missing the broader context of the action. This shortcoming highlights a key limitation of current video-based MAEs in action recognition: while they learn spatiotemporal representations, they may focus on isolated spatial details rather than capturing the temporal coherence needed to understand complex actions. Consequently, these models may excel in detecting spatial features but struggle to fully capture the sequential dynamics critical for action-specific contexts.

Encouraged by the intuition of how humans integrate minor scene changes naturally over short video sequences (5–10 seconds) without losing context, which is crucial for interpreting dynamic visual content. Cognitive neuroscience research highlights that integrating spatiotemporal information from videos with spatial information from a sequence of frames (from here onward this is referred to as 'spatiotemporal-spatial' unless otherwise stated) is a first capability developed early in infancy~\cite{richards2003development,van1996spatiotemporal}, often before semantic traits are learned~\cite{henderson2003human}. Later, Human visual processing semantically relates action cues across frames, forming a cohesive, action-focused understanding of video content. Inspired by this, we propose \textit{CrossVideoMAE} a cross-modal contrastive learning framework to enhance spatiotemporal and semantic representation learning for video understanding. CrossVideoMAE leverages Masked Image Modeling (MIM)~\cite{bao2022beit, he2022masked} for video to capture correlations between frames and their temporal contexts, integrating semantic attributes without relying on costly language annotations, as evidenced by VARD~\cite{lin2023self}. Through contextual cues within frames, models can detect patterns, objects, interactions, and transitions, achieving high-level semantic understanding even without explicit language data. Temporal relationships among sampled frames further enhance this by learning motion patterns and interactions critical for understanding video dynamics.

CrossVideoMAE extracts mutually correlated spatiotemporal and semantic attributes from videos and sampled frames in a single, end-to-end pre-training phase. Static scene attributes from sampled frames (e.g., "basketball," "hands/arms", "face") complement general action attributes from videos (e.g., "shooting," "trajectory", "arm movement"), enhancing video comprehension. While directly distilling semantic features from state-of-the-art image-based MAEs~\cite{he2022masked, li2022semmae} into video-based models is theoretically possible, significant challenges arise. Unlike image-based MAEs focused on spatial semantics, video-based MAEs must handle temporal complexity, capturing motion, interaction, and continuity. This task is inherently complex, as video-based models must learn from frame sequences, which introduces temporal redundancy and correlation issues absent in static images.

CrossVideoMAE addresses these challenges by independently learning spatial and temporal representations and refining their integration to enhance video content comprehension. Unlike CrossVideo~\cite{liu2024crossvideo}, which emphasizes cross-modal learning, CrossVideoMAE specifically enhances spatiotemporal modelling within a Vision Transformer (ViT)-based MAE framework. To the best of our knowledge, this is the first study to leverage a pre-sampled dataset for SSL in video learning. CrossVideoMAE encodes spatiotemporal-spatial feature correspondences between video sequences and sampled frames, using pre-trained MAEs to embed visible tokens from raw videos, augmented sequences, and frames in a unified feature space with varied masking ratios. This approach enforces spatiotemporal consistency between the two modalities, enabling robust video encoders that capture visual-semantic invariant representations transferable to downstream tasks. Our contributions can be summarized as follows:

\begin{itemize}
\item We show that self-supervised contrastive learning effectively captures spatiotemporal-spatial feature correspondence by enforcing human-like priors on learned concepts, relating video tokens to sampled frames with varied masking.

\item We propose an end-to-end self-supervised contrastive framework that aligns video and frame embeddings, ensuring invariance to augmentations.

\item Our method embeds visible tokens at video and frame levels, capturing correlations and enhancing temporal dynamics understanding.

\item Extensive experiments demonstrate that our approach achieves competitive performance with significantly lower computational resources than existing methods.

\end{itemize}

\section{Related Work}
\label{sec:related}

\noindent \textbf{Representation Learning on Videos:}
SSL video representation learning has been extensively studied, with early work leveraging pretext tasks like temporal order, space-time puzzles, and optical flow statistics for supervision~\cite{benaim2020speednet,misra2016shuffle,wang2015unsupervised,xu2019self}. Recently, contrastive learning has gained prominence by enforcing feature space invariance, bringing positive samples closer and separating negatives~\cite{chen2020simple,he2020momentum,li2021improve}. Several video-based methods have extended this by exploring spatial-temporal augmentations~\cite{diba2021vi2clr,feichtenhofer2021large,ge2021revitalizing,han2020memory,pan2021videomoco,qian2021spatiotemporal}. Additionally, Masked Image Modeling (MIM)~\cite{bao2022beit,he2022masked,feichtenhofer2022masked,wei2022masked} has been successfully adapted to videos~\cite{feichtenhofer2022masked,tong2022videomae,wei2022masked}, achieving strong results across various video tasks~\cite{simonyan2014two,tran2015learning,wang2018temporal,wang2021tdn,zhang2021co}.

\noindent \textbf{Masked Autoencoders (MAEs):}
MAEs~\cite{bandara2023adamae,feichtenhofer2022masked,huang2023mgmae,tong2022videomae,wang2023videomae} have made significant advances over contrastive learning in self-supervised vision tasks by utilizing high masking ratios during pre-training, resulting in simpler, more efficient models. Masking techniques are central to their success~\cite{feichtenhofer2022masked,tong2022videomae}, with common strategies including patch masking~\cite{feichtenhofer2022masked}, frame masking~\cite{qian2021spatiotemporal,wei2022masked}, and tube-based masking, which drops tokens across frames to avoid information leakage~\cite{wang2023videomae}. However, no single masking method generalizes well across datasets due to varying scene dynamics, data acquisition conditions, and spatiotemporal complexities~\cite{bandara2023adamae}. For instance, SpatioTemporalMAE~\cite{feichtenhofer2022masked} excelled on Kinetics-400 with random patch masking, while VideoMAE~\cite{tong2022videomae} performed best on Something-Something V2 using tube masking, highlighting the need for task-specific masking strategies.

\noindent \textbf{MAEs for Videos:}
Extending MAEs to videos, SpatioTemporalMAE~\cite{feichtenhofer2022masked} and VideoMAE~\cite{tong2022videomae} have made notable progress. BEVT~\cite{wang2022bevt} and OmniMAE~\cite{girdhar2023omnimae} further advanced the field by training unified image and video MAEs with shared weights across datasets. MAR~\cite{qing2023mar} reduced computational costs by using running cell masking, while VideoMAE v2~\cite{wang2023videomae} proposed masking decoder-reconstructed tokens. AdaMAE~\cite{bandara2023adamae} introduced adaptive masking to replace random techniques. Human priors, such as motion trajectories, were incorporated in MGMAE~\cite{huang2023mgmae}, MotionFormer~\cite{patrick2021keeping}, and MME~\cite{sun2023masked}, while SemMAE~\cite{li2022semmae} used semantic parts-guided masking. MaskViT~\cite{gupta2022maskvit} added spatial and spatiotemporal attention with variable token masking ratios.

\noindent \textbf{Cross-Modal Representation Learning:}Videos often include multiple modalities such as text, images, motion (e.g., optical flow), and audio, which provide rich supervision for understanding semantic context~\cite{desai2021virtex,radford2021learning,sariyildiz2020learning,castrejon2016learning,gong2014improving,karpathy2015deep,lu202012,miech2020end}. Cross-modal pre-training, combining text with images~\cite{desai2021virtex,radford2021learning} and audio with video~\cite{arandjelovic2017look,arandjelovic2018objects,morgado2021robust,morgado2021audio,korbar2018cooperative,owens2018audio}, has shown success in learning transferable representations for various downstream tasks. Approaches like BEVT~\cite{wang2022bevt} and OmniMAE~\cite{girdhar2023omnimae} integrate image and video pre-training, while CrossVideo~\cite{liu2024crossvideo} introduces point cloud video datasets paired with image datasets. Our method, however, addresses the lack of pre-sampled frame datasets for images by introducing a new sampling strategy for the image branch. We manually sample frames to enhance learning since video frames provide richer semantic context. In this context, CrossVideoMAE fuses SpatioTemporalMAE~\cite{feichtenhofer2022masked} (video branch) with a pre-trained MAE~\cite{he2022masked} (image branch) using ViT-B/16. This method aligns feature embeddings from sampled frames with corresponding videos at both frame and video levels, ensuring robustness to video augmentations while distilling semantic knowledge effectively.

\section{Proposed Method}
\label{sec:method}

\begin{figure*}[h!]
\vspace{-15pt}
\centering
\includegraphics[width=\textwidth]{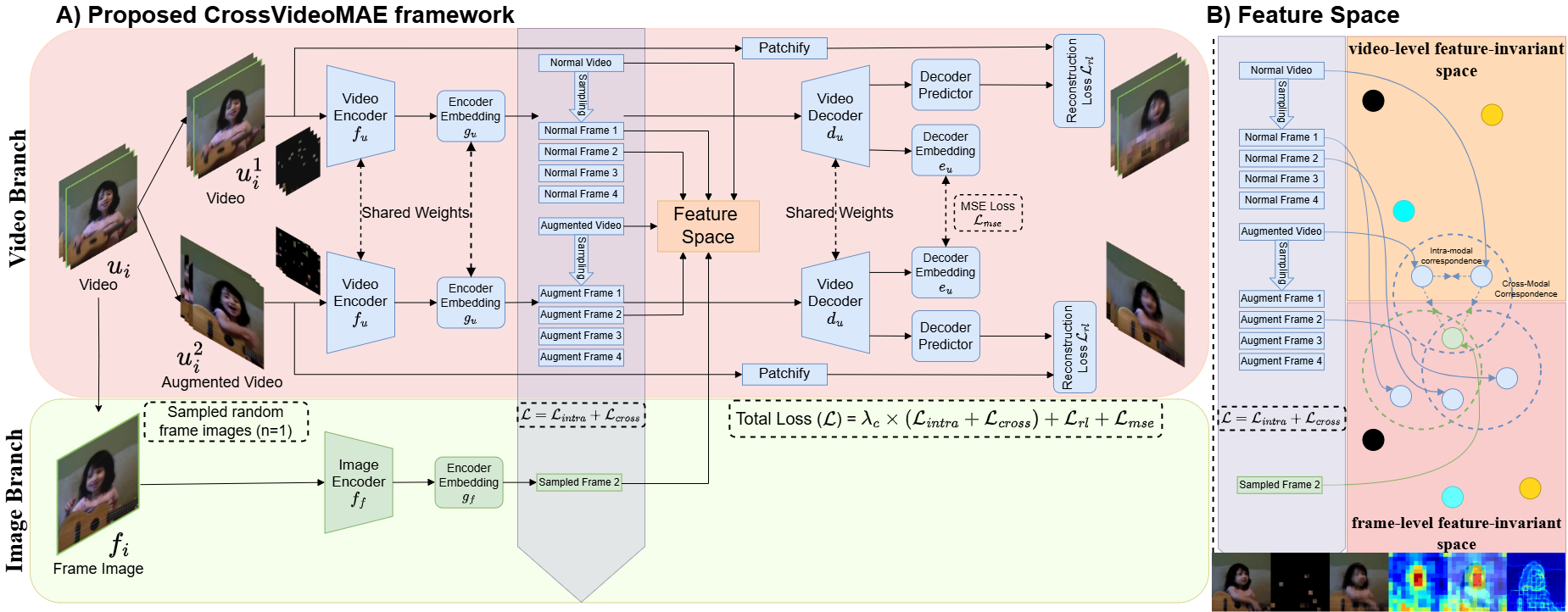}
\vspace{-18pt}
\captionsetup{font=footnotesize}
\caption{ \textbf{A).} The proposed CrossVideoMAE framework comprises two branches: the video branch and the image branch. The video branch employs intra-modal pre-training to ensure that the encoder develops invariance to augmentations within the video domain. The image branch leverages are cross-modal pre-training to distill semantic knowledge from pre-trained MAE~\cite{he2022masked}, transferring insights from sampled frames to corresponding videos. The model is pre-trained jointly across video and image domains using a combination of intra-modal and cross-modal contrastive learning objectives at both the video and frame levels. For downstream tasks, the image branch is discarded, and only the video branch encoder is utilized as the backbone. \textbf{B).} Zoom in version of the feature space. This approach demonstrates the spatiotemporal-spatial alignment of feature embedding correspondence for visible tokens, ensuring invariance at both the video level and frame level, enhancing the representation robustness.}
\label{fig:architecture}
\vspace{-15pt}
\end{figure*}

The overall architecture of our proposed method is illustrated in Fig.~\ref{fig:architecture}. In this section, we enhance self-supervised video representation learning by integrating both intra-modal and cross-modal contrastive learning at both video and frame levels. We provide a detailed explanation of our approach by adapting the design and methods described in ~\cite{liu2024crossvideo}. We begin by outlining the network architecture of the proposed method in \S~\ref{sec:perliminaries}. Subsequently, in \S~\ref{sec:intra-modal} and \S~\ref{sec:cross-modal}, we describe the contrastive learning loss functions developed for intra-modal and cross-modal settings at both video and frame levels. Finally, we detail our overall pre-training objective in \S~\ref{sec:objective}.

\subsection{Preliminaries}
\label{sec:perliminaries}
\textbf{Problem Setup:} Suppose that there is a dataset provided \(\mathcal{D} = \{(u_i, f_i)\}_{i=1}^{|\mathcal{D}|}\), with \(u_i \in \mathbb{R}^{\textcolor{blue}{T} \times \textcolor{teal}{H} \times \textcolor{teal}{W} \times \textcolor{red}{C}}\) and \(f_i \in \mathbb{R}^{\textcolor{teal}{H} \times \textcolor{teal}{W} \times \textcolor{red}{C}}\). Note that $f_i$ is obtained by randomly sampling frames from the video sequence $u_i$, where $u_i$ has a temporal sequence of frames of length $\textcolor{blue}{T}$. We define each $u_i$ as $u_i = \{f_1, f_2, \ldots, f_j, \ldots, f_{\textcolor{blue}{T}}\}$, where each $f$ denotes one frame. We tokenize the video and sampled frames into a sequence of tokens \(u_i = \{u_i^1, u_i^2, \ldots, u_i^N\}\), and \(f_i = \{f_i^1, f_i^2, \ldots, f_i^M\}\) for each sample $i$. For its masked version, we denote the visible tokens as $\{u_i^v\}, \{f_i^v\}$. The feature embedding of visible tokens $\{u_i^v\}$ obtained by $f_{u}(\{u_i^v\} + \{p_i^v\})$, where $\{p_i^v\}$ is the positional encoding. Our goal is to pre-train a video encoder \(f_{u}(\cdot)\) in a self-supervised fashion to be effectively transferable to downstream tasks. To this end, we use an image encoder \(f_{f}(\cdot)\), encoder embedding with multi-layer perceptron (MLP) \(g_{u}(\cdot)\) and \(g_{f}(\cdot)\) for the video and image, respectively. 
\textbf{Notations:} $u_i: \{u_i^v\} + \{p_i^v\}$,
$f_i: \{f_i^v\} + \{p_i^v\}$

\subsection{Intra-Modal Contrastive Learning}
\label{sec:intra-modal}
Building on the success of contrastive learning in image and video domains, we posit that intra-modal contrastive learning is essential for capturing view-invariant representations. At both video and frame levels, we enforce the feature embeddings of visible tokens to be invariant to a variety of data augmentations. For a given input video \(u_i\), we denote the visible tokens of the raw and augmented video as \(u_{i}^{1}\) and \(u_{i}^{2}\), respectively. The augmented video \(u_{i}^{2}\) is constructed by sequentially applying spatiotemporal augmentations to the original video and then randomly masking portions of the augmented video. These augmentations include transformations such as rotation, random cropping, scaling, and translation. Additionally, we apply spatial transformations like colour jittering, spatiotemporal transformations such as random augmentation, random resizing, cropping, horizontal flipping, random erasing, mixup, and cut mix, along with temporal transformations like frame extraction through down-sampling.

The video encoder \(f_{u}\) maps both \(u_{i}^{1}\) and \(u_{i}^{2}\) into a feature embedding space. These embedded vectors are then projected into a video-level invariant space \(\mathbb{R}^{|\mathcal{D}|}\) using the encoder embedding function \(g_{u}(\cdot)\). Subsequently, these projected vectors in the video-level invariant space $\mathbb{R}^{|\mathcal{D}|}$ are sampled to obtain frame-level invariant space within the same embedding space \(\mathbb{R}^{|\mathcal{D}|}\), where the contrastive loss is applied. This sampling process involves extracting frame-level embedding corresponding to each frame from video-level feature embedding of visible tokens to capture temporal variations effectively. Sampling from the video-level invariant space \(\mathbb{R}^{|\mathcal{D}|}\) to obtain frame-level invariant space involves extracting frame-specific embeddings from video-level feature embedding of the visible tokens. These frame-level embeddings correspond to each frame within the same embedding space \(\mathbb{R}^{|\mathcal{D}|}\), as in MAEs frame-level embedding distinguishable within the video-level embedding. We denote the video-level projected vectors of \(u_{i}^{1}\) and \(u_{i}^{2}\) as \(\textbf{z}_{u_i}^{1}\) and \(\textbf{z}_{u_i}^{2}\), respectively, and the frame-level projected vectors as \(\textbf{z}_{f_i}^{1}\) and \(\textbf{z}_{f_i}^{2}\). Here, each projected vector \(\textbf{z}_{i}^t\) is defined as \(\textbf{z}_{i}^t = g_{u}(f_{u}(u_{i}^t))\). The frame-level projected vectors $\textbf{z}_{f_i}^t$ is obtained manually by sampling the video-level projected vectors $\textbf{z}_{u_i}^t$. 

For both the video-level and frame-level objectives, our aim is to maximize the cosine similarity between \(\textbf{z}_{u_i}^{1}\) and \(\textbf{z}_{u_i}^{2}\) for video-level learning, and between \(\textbf{z}_{f_i}^{1}\) and \(\textbf{z}_{f_i}^{2}\) for frame-level learning, while minimizing the similarity with all other projected vectors within the mini-batch. We utilize the NT-Xent loss, as introduced in SimCLR~\cite{chen2020simple}, is used, to learn discriminative features. Notably, our approach does not rely on any memory bank, consistent with recent advancements in self-supervised contrastive learning. For both video and frame levels, we compute the loss functions \(L_{u}(i, 1, 2)\) and \(L_{f}(i, 1, 2)\) for the positive pairs \(\textbf{z}_{u_i}^{1}\) and \(\textbf{z}_{u_i}^{2}\), and \(\textbf{z}_{f_i}^{1}\) and \(\textbf{z}_{f_i}^{2}\), respectively, as follows:

\begin{equation}
\scriptsize
L_{u}(i, 1, 2) = -\log \frac{\exp(\text{s}(\textbf{z}_{u_i}^{1}, \textbf{z}_{u_i}^{2})/\tau)}{\sum\limits_{\substack{k=1 \\ k \neq i}}^{N} \exp(\text{s}(\textbf{z}_{u_i}^{1}, \textbf{z}_{u_k}^{1})/\tau) + \sum\limits_{k=1}^{N}\exp(\text{s}(\textbf{z}_{u_i}^{1}, \textbf{z}_{u_k}^{2})/\tau)}
\label{equ:equation1}
\end{equation}
\begin{equation}
\scriptsize
L_{f}(i, 1, 2) = -\log \frac{\exp(\text{s}(\textbf{z}_{f_i}^{1}, \textbf{z}_{f_i}^{2})/\tau)}{\sum\limits_{\substack{k=1 \\ k \neq i}}^{N} \exp(\text{s}(\textbf{z}_{f_i}^{1}, \textbf{z}_{f_k}^{1})/\tau) + \sum\limits_{k=1}^{N}\exp(\text{s}(\textbf{z}_{f_i}^{1}, \textbf{z}_{f_k}^{2})/\tau)}
\label{equ:equation2}
\end{equation}
\noindent where N is the mini-batch size, $\tau$ is the temperature coefficient and \(\text{s}(\cdot)\) denotes the cosine similarity function. Our intra-modal instance discrimination contrastive loss function $\mathcal{L}_{\text{\textit{intra}}}$ for a mini-batch can be described as:
\vspace{-5pt}
\begin{equation}
\mathcal{L}_{\text{\textit{intra}}} = \frac{1}{2N}{\sum_{i=1}^{N} (L_u(i, 1, 2) + L_f(i, 1, 2)})
\end{equation}
\vspace{-5pt}

\subsection{Cross-Modal Contrastive learning}
\label{sec:cross-modal}
In addition to aligning feature embeddings within the video domain (intra-modal contrastive learning), we introduce an auxiliary cross-modal contrastive learning objective that spans both video and sampled frame image modalities. This approach is designed to learn discriminative features across modalities and enhance the video encoder's ability, thereby improving the representation learning capability for videos by aligning frame-level features with their corresponding sampled frame image features. Specifically, we first embed the visible tokens of sampled frames \(f_i\) of \(u_i\) into a feature embedding space using the image encoder \(f_{f}(\cdot)\). We then project the embedded vectors into the frame-level feature invariant space$\mathbb{R}^{|\mathcal{D}|}$ using the image encoder embedding \(g_{f}(\cdot)\), defined as \(\textbf{h}_{f_i}\) where $\textbf{h}_i = g_{f}(f_{f}(f_i))$. The difference between the frame-level representation \(\textbf{h}_{f_i} \) in the cross-modal section and intra-modal section lies in the masking ratio applied to each branch. In contrast to previous cross-modal approaches, we do not explicitly discriminate features between the two modalities (video and image). Instead, we implement feature discrimination in the video domain and distill semantic attributes from sampled frames to videos to improve video understanding. Then, we compute the mean of the projected vectors $\textbf{z}_{u_i}^{1}$ and $\textbf{z}_{u_i}^{2}$, $\textbf{z}_{f_i}^{1}$ and $\textbf{z}_{f_i}^{2}$ to obtain the projected at video-level and frame-level features $\textbf{z}_{u_i}$ and $\textbf{z}_{f_i}$ of $u_i$. 
\vspace{-3pt}
\begin{equation}
\textbf{z}_{u_i} = \frac{1}{2} (\textbf{z}_{u_i}^{1} + \textbf{z}_{u_i}^{2}) ; \quad \textbf{z}_{f_i} = \frac{1}{2} (\textbf{z}_{f_i}^{1} + \textbf{z}_{f_i}^{2})
\end{equation}
In the invariant space, our goal is to maximize the cosine similarity between \(\textbf{z}_{f_i}\) and \(\textbf{h}_{f_i}\), as well as between \(\textbf{z}_{u_i}\) and \(\textbf{h}_{u_i}\), since they correspond to the same instance. Our cross-modal alignment strategy compels the model to learn from more challenging positive and negative samples, thereby enhancing the representation capability beyond what is achieved through intra-modal alignment alone. We compute the contrastive loss functions \(C_u(i, 1, 2)\) and \(C_f(i, 1, 2)\) for the positive pairs \(\textbf{z}_{u_i}\) and \(\textbf{h}_{u_i}\) with \(\textbf{z}_{f_i}\) and \(\textbf{h}_{f_i}\) as follows:
\begin{equation}
\scriptsize
C_u(i, 1, 2) = -\log \frac{\exp(\text{s}(\textbf{z}_{u_i}, \textbf{h}_{u_i})/\tau)}{\sum\limits_{\substack{k=1 \\ k \neq i}}^{N} \exp(\text{s}(\textbf{z}_{u_i}, \textbf{z}_{u_k})/\tau) + \sum\limits_{k=1}^{N}\exp(\text{s}(\textbf{z}_{u_i}, \textbf{h}_{u_k})/\tau)}
\end{equation}
\begin{equation}
\scriptsize
C_f(i, 1, 2) = -\log \frac{\exp(\text{s}(\textbf{z}_{f_i}, \textbf{h}_{f_i})/\tau)}{\sum\limits_{\substack{k=1 \\ k \neq i}}^{N} \exp(\text{s}(\textbf{z}_{f_i}, \textbf{z}_{f_k})/\tau) + \sum\limits_{k=1}^{N}\exp(\text{s}(\textbf{z}_{f_i}, \textbf{h}_{f_k})/\tau)}
\end{equation}
\noindent where \(\text{s}(\cdot)\), N, $\tau$ refers to the same parameters as in Eq.~\ref{equ:equation1},~\ref{equ:equation2} The cross-modal loss function $\mathcal{L}_{\text{\textit{cross}}}$ for a mini-batch is then formulated as:
\vspace{-5pt}
\begin{equation}
\mathcal{L}_{\text{\textit{cross}}} = \frac{1}{2N}{\sum_{k=1}^{N} (C_u(i, 1, 2) + C_f(i, 1, 2)})
\end{equation}

The difference between the frame-level representation in the cross-modal section and the intra-modal section lies in the difference between the masking ratio applied to each branch,with each capturing different aspects of information in each context: cross-modal branch focusing on alignment across modalities and the intra-modal branch focusing on temporal consistency.

\subsection{MSE and Reconstruction Losses}

The MSE and reconstruction losses are tangential to the cross-modal contrastive learning. While the contrastive loss aligns features within and across modalities, the MSE loss focuses specifically on reconstruction fidelity, which helps the model retain finer details in the decoded output and improve the feature embedding of visible tokens in the invariant-space and indirectly contributes to reducing the contrastive loss, and the reconstruction loss on both the original and augmented videos helps the model to generate better feature embedding of visible tokens, which enables more accurate reconstruction of original video and augmented video. With improved feature embedding of visible tokens intra-modal contrastive learning further promotes invariance to augmentations, allowing the model to maintain consistency across augmented views. 

\noindent \textbf{MSE Loss:} The MSE loss is applied at the decoder side to ensure accurate reconstruction of both the original and augmented videos, while minimizing the distance between them. Given the two representations \(f_{e, 1}\) and \(f_{e, 2}\), where \(f_{e, 1} = f_{u}(u_i^{1})\), \(f_{e, 2} = f_{u}(u_i^{2})\), after the decoder attention block, we will obtain two predicted representations, \(f_{d, 1}\) and \(f_{d, 2}\). Then, the decoder prediction is applied between them to get the MSE loss, defined as:
\begin{equation}
\mathcal{L}_{mse} = \frac{1}{N} \sum_{k=1}^{N} (\mathcal{L}_{\textit{pred}}(f_{d, 1}, f_{d, 2}))
\end{equation}
\noindent where N is the batch size and the prediction loss $\ell_{\text{\textit{pred}}}$ is defined as:
\begin{equation}
\mathcal{L}_{\textit{pred}}(f_{d, 1}, f_{d, 2}) = \left\| e_{u}(f_{d, 1}) - e_{u}(f_{d, 2}) \right\|_2^2
\end{equation}

\noindent where the decoder embedding, \(e_{u}(\cdot)\), is a MLP. 

\noindent \textbf{Reconstruction Loss:} The reconstruction loss is calculated using the Mean Squared Error (MSE) between the decoder predicted and target representations. In addition to the MSE (decoder embedding) loss, the decoder performs reconstruction for both the original video and the augmented video. Therefore, the reconstruction loss is defined as:
\begin{equation}
\mathcal{L}_{rl} = \frac{1}{N} \sum_{k=1}^{N} (\left\| u_{i, 1} - \tilde{u}_{i, 1} \right\|_2^2 + \left\| u_{i, 2} - \tilde{u}_{i, 2} \right\|_2^2)
\end{equation}
where N is the batch size, $\tilde{u}_{i, 1}$ and $\tilde{u}_{i, 2}$ are predicted representations.

\subsection{Overall Objective}
\label{sec:objective}

Finally, the overall loss function during pre-training is derived as a combination of contrastive loss c (multiplied by a weight $\lambda_c$), reconstruction loss, and MSE loss. The \(\mathcal{L}_{intra}\) loss ensures invariance to spatiotemporal augmentations, while the \(\mathcal{L}_{cross}\) loss maintains spatiotemporal-spatial correspondence and distills rich semantic information from sampled frames to videos. Additionally, the \(\mathcal{L}_{rl}\) and \(\mathcal{L}_{mse}\) losses are employed to enforce the model to reconstruct the data while preserving intricate relationships within the input. Together, these loss components contribute to a robust and comprehensive pre-training objective, enhancing the model's ability to learn meaningful and discriminative video representations.
\vspace{-5pt}
\begin{equation}
\mathcal{L} = \lambda_c \times (\mathcal{L}_{intra} + \mathcal{L}_{cross}) + \mathcal{L}_{rl} + \mathcal{L}_{mse} 
\end{equation}
\vspace{-5pt}

\begin{table*}[t!]
\vspace{-20pt}
\centering
\scriptsize
\begin{tabular}[t]{lccccccc}
\toprule
\multirow{2}{*}{\textbf{Method}} & \multirow{2}{*}{\textbf{Backbone}} & \textbf{Extra pre-trainining} & \textbf{Param} & \multicolumn{4}{c}{\textbf{Action Full Fine-tuning (Acc@1 (\%))}} \\
& & \textbf{dataset} & \textbf{(M)} & \textbf{UCF101}~\cite{soomro2012ucf101} & \textbf{HMDB51}~\cite{kuehne2011hmdb} & \textbf{SSv2}~\cite{goyal2017something} & \textbf{K400}~\cite{kay2017kinetics} \\
\midrule\midrule
SpeedNet~\cite{benaim2020speednet} & S3D-G & K400 & 9 & 81.1 & 48.8 & — & — \\
Pace Pred~\cite{wang2020self} & R(2+1)D & K400 & 15 & 77.1 & 36.6 & — & — \\
Vi$^2$CLR~\cite{diba2021vi2clr} & S3D & UCF101 & 9 & 82.8 & 52.9 & — & — \\
Vi$^2$CLR~\cite{diba2021vi2clr} & S3D & K400 & 9 & 89.1 & 55.7 & — & — \\
MemDPC~\cite{han2020memory} & R2D3D-34 & K400 & 32 & 86.1 & 54.5 & — & — \\
RSPNet~\cite{chen2021rspnet} & R(2+1)D & K400 & 9 & 81.1 & 44.6 & — & — \\
RSPNet~\cite{chen2021rspnet} & S3D-G & K400 & 9 & 93.7 & 64.7 & — & — \\
VideoMoCo~\cite{pan2021videomoco} & R(2+1)D & K400 & 15 & 78.7 & 49.2 & — & — \\
HiCo~\cite{qing2022learning} & S3D-G & UK400 & — & 91.0 & 66.5 & — & — \\
CVRL~\cite{qian2021spatiotemporal} & SlowOnly-R50 & K400 & 32 & 92.9 & 67.9 & — & — \\
CVRL~\cite{qian2021spatiotemporal} & SlowOnly-R152 & K600 & 32 & 94.4 & 70.6& — & — \\

MIL-NCE~\cite{miech2020end} & S3D-G & HowTo100M & 9 & 91.3 & 61.0 & — & — \\
MMV~\cite{alayrac2020self} & S3D-G & AS+HTM & 9 & 92.5 & 69.6 & — & — \\
CPD~\cite{li2020learning} & ResNet50 & IG300k & — & 92.8 & 63.8 & — & — \\
ELO~\cite{piergiovanni2020evolving} & R(2+1)D & Youtube8M-2 & — & 93.8 & 67.4 & — & — \\
XDC~\cite{alwassel2020self} & R(2+1)D & IG65M & 15 & 94.2 & 67.1 & — & — \\
GDT~\cite{patrick2021multi} & R(2+1)D & IG65M & 15 & 95.2 & 72.8 & — & — \\

\midrule

\multicolumn{6}{l}{\textbf{\textit{Pre-trained Epochs:}}} & \textbf{\textit{800}} & \textbf{\textit{1600}} \\
\midrule\midrule
VIMPAC~\cite{tan2021vimpac} & ViT-L & HowTo100M+DALLE & 307 & 92.7 & 65.9 & 68.1 & 77.4 \\
SVT~\cite{ranasinghe2022self} & ViT-B &IN-21K+K400 & 121 & 93.7 & 67.2 & — & — \\
BEVT~\cite{wang2022bevt} & Swin-B & IN-1K+K400+DALLE & 88 & — & — & 70.6 & 80.6 \\
SpatioTemporalMAE~\cite{feichtenhofer2022masked} & ViT-B & — & 87 & — & — & 68.3 & 81.3 \\
MME~\cite{sun2023masked} & ViT-B & — & 87 & 96.5 & \textcolor{blue}{78.0} & 70.0 & 81.8 \\
VideoMAE~\cite{tong2022videomae} & ViT-B & — & 87 & 90.8 & 61.1 & — & — \\
MAR~\cite{qing2023mar} & ViT-B & — & 87 & 91.0 & 61.4 & — & — \\
\midrule
& & & & \multicolumn{2}{c}{\textbf{\textit{Pre-trained Model: K400}}} & & \\
\midrule
VideoMAE~\cite{tong2022videomae} & ViT-B & — & 87 & 96.1 & 73.3 & 69.3 & 80.9 \\
MAR~\cite{qing2023mar} & ViT-B & — & 87 & 95.9 & 74.1 & 71.0 & 81.0 \\
\midrule
OmniMAE~\cite{girdhar2023omnimae} & ViT-B & IN-1K & 87 & — & — & 69.3 & 80.6 \\
ViC-MAE~\cite{hernandez2023visual} (T.L.) & ViT-B & K(4,6,7)+MiT+IN-1K & 87 & — & — & 69.8 & 80.9 \\
ConvMAE~\cite{gao2022mcmae} & ConvViT-B & — & 86 & — & — & 69.9 & 81.7 \\
AdaMAE~\cite{bandara2023adamae} & ViT-B & — & 87 & — & — & 70.0 & 81.7 \\
MGMAE~\cite{huang2023mgmae} & ViT-B & — & 87 & — & — & 70.6 & 81.2 \\
CMAE-V~\cite{lu2023cmae} & ConvViT-B & — & 87 & — & — & 71.1 & 82.2 \\
MVD-B~\cite{wang2023masked} & Teacher-B & IN-1K & 87 & \textcolor{blue}{97.0} & 76.4 & \textcolor{blue}{72.5} & \textcolor{blue}{82.7} \\
\rowcolor[gray]{0.9}
\textbf{CrossVideoMAE} & \textbf{ViT-B} & \textbf{IN-1K} & \textbf{87*} & \textcolor{red}{97.6} & \textcolor{red}{78.4} & \textcolor{red}{73.7} & \textcolor{red}{83.2} \\
\bottomrule
\end{tabular}
\vspace{-5pt}
\captionsetup{font=footnotesize}
\caption{Comparison of our proposed method with state-of-the-art supervised (ref to supplementary material) and self-supervised methods on the UCF101, HMDB51, K400, and SSv2 dataset using the ViT-B/16 backbone. The best results are highlighted in \textcolor{red}{red}, and the second-best results in \textcolor{blue}{blue}. T.L: Transfer Learning. IN: ImageNet dataset. K(4,6,7): Kinetics-400, 600, and 700 datasets. *: shared parameters.}
\label{tab:results}
\vspace{-15pt}
\end{table*}

\section{Experiments}
\label{sec:experiments}

\subsection{Datasets}

We evaluated our method on four action recognition video datasets: UCF101~\cite{soomro2012ucf101}, HMDB51~\cite{kuehne2011hmdb}, Kinetics-400 (K400)~\cite{kay2017kinetics}, and Something-Something V2 (SSv2)~\cite{goyal2017something} (\textit{refer to supplementary materials for details}).

For K400 and SSv2, where pre-sampled frame image datasets were unavailable, we created smaller subsets of the original datasets. This approach maintained class diversity while addressing the challenges of manual frame extraction from large datasets, which can be resource-intensive due to GPU constraints. We randomly sampled frames from each video in these smaller datasets to construct corresponding pre-sampled frame datasets. Ablation studies were performed on SSv2, and the optimized parameters were used for both K400 and SSv2 evaluations. For UCF101 and HMDB51, we fine-tuned a pre-trained K400 model directly without additional sampling.

\subsection{Preprocessing}

We follow the preprocessing protocols outlined in MAE~\cite{he2022masked} and SpatioTemporalMAE~\cite{feichtenhofer2022masked}. For the Kinetics-400 and SSv2 datasets, we sample 16 frames, from each raw video, setting each frame at a resolution of 224 × 224 pixels, and apply a temporal sliding window with a stride of 4, with the starting frame location selected randomly~\cite{feichtenhofer2022masked}. The default resolution for the videos and their corresponding sampled frames is 224 × 224. In addition to spatiotemporal augmentations such as random augmentation, random resizing, cropping,
horizontal flipping, random erasing mixup, and cutmix applied in the video branch (\S~\ref{sec:intra-modal}), we also apply spatial augmentations random resizing, cropping, and horizontal flipping to the sampled frames.

\subsection{Implementation Details}

We designed a three-tower architecture inspired by recent advancements in Masked Autoencoders (MAEs)~\cite{he2022masked,tong2022videomae,feichtenhofer2022masked}, as shown in Figure~\ref{fig:architecture}. The network comprises two main branches: the video branch and the image branch.

\noindent\textbf{Video Branch:} This branch comprises two shared-weight pre-trained SpatioTemporalMAE~\cite{feichtenhofer2022masked} configurations based on ViT-B/16, with masking ratios (\(\rho\)) ranging from 90\% to 95\%. These models take as input both the original and an augmented version of the video. The video encoder extracts video-level features and samples them to generate frame-level feature embeddings of the visible tokens. This encoder embedding facilitates improved information exchange across a sequence of frames, effectively capturing spatiotemporal dynamics.

\noindent \textbf{Image Branch:} The Image branch is built on a pre-trained MAE~\cite{he2022masked} ViT-B/16 configuration with a masking ratio (\(\rho\)) between 75\% and 90\%. This branch extracts frame-level feature embeddings of visible tokens, leveraging the spatial priors learned on sampled frames. These priors, which can be considered as a form of human prior, assist in learning spatial information for each frame in the video sequence. The encoder embedding further distils semantic knowledge from these sampled frames to the videos, enhancing the overall understanding of spatiotemporal content.

We utilize the test-time adaptation technique to mitigate the need for a large number of GPU resources to save GPU memory and reduce pre-training time. We use a patch size of \textcolor{blue}{2} $\times$ \textcolor{red}{3} $\times$ \textcolor{teal}{16} $\times$ \textcolor{teal}{16}, resulting in $\textcolor{blue}{\left(\frac{16}{2}\right)} \times \textcolor{red}{\left(\frac{3}{3}\right)} \times \textcolor{teal}{\left(\frac{224}{16}\right)} \times \textcolor{teal}{\left(\frac{224}{16}\right)} = \textcolor{purple}{1568}$ tokens for an input video of size \textcolor{blue}{16} $\times$ \textcolor{red}{3} $\times$ \textcolor{teal}{224} $\times$ \textcolor{teal}{224}. The differences in masking ratios and spatiotemporal-spatial feature correspondence strengthen our method. Both pre-trained SpatioTemporalMAE~\cite{feichtenhofer2022masked} and MAE~\cite{he2022masked} allow higher masking ratio ($\textcolor{orange}{\rho}$) and distill well-learned semantic attributes from sampled frames to videos through the difference between masking ratios, and spatiotemporal-spatial feature embedding correspondence of visible tokens. For all experiments, We use adamW~\cite{loshchilov2018decoupled} optimizer with a batch size of 32 and 8 GPUs with a decoder depth of 4. 

\subsection{Test-Time Adaptation (TTA)} 
Initial inference is conducted using pre-trained weights available from open-source repositories (MAE ~\cite{he2022masked} for the image branch and SpatioTemporalMAE ~\cite{feichtenhofer2022masked} for the video branch) to compute losses. We perform 20 gradient updates based on these losses during test time, refining the model weights. After the adaptation step, the final inference is performed to obtain the refined weights. This approach saves GPU memory and reduces pre-training time. TTA is a refinement step, not a replacement for pre-training, dynamically fine-tuning pre-trained weights (from MAE and SpatioTemporalMAE) to better align with test-time data during inference. Due to GPU constraints, we created smaller subsets of K400 and SSv2 from the subset randomly sampled frames to create corresponding pre-sampled frame datasets-our test-time data, as manual frame extraction from large datasets is resource-intensive. TTA complements pre-training by efficiently adapting weights with 20 lightweight gradient updates per batch, based on contrastive and reconstruction losses on test-time data, without requiring large-scale re-training.

\section{Results}
\label{sec:results}

We evaluate the performance of CrossVideoMAE in action recognition through end-to-end fine-tuning, following established protocols. For the SSv2 and K400 datasets, we apply the methodologies used in previous works~\cite{bao2022beit,he2022masked,feichtenhofer2022masked}, while for UCF101 and HMDB51, we adopt the protocols proposed by Ranasinghe et al.~\cite{ranasinghe2022self}.

\subsection{Comparison with State-of-the-Art Methods}
We compared our method with SOTA video SSL action recognition models on the UCF101, HMDB51, K400, and SSv2 datasets under the full fine-tuning setting (Tab.~\ref{tab:results}). Linear classification results, comparisons with supervised models on K400 and SSv2, and video retrieval results are provided in the supplementary material. Our approach utilizes the ViT-B/16 architecture, with approximately 87 million shared parameters. For inference, we employed multiview testing with \( K \) temporal clips (K = 2 for SSv2 and K = 7 for K400) and 3 spatial views per clip, averaging the results across all views for the final prediction.

CrossVideoMAE consistently outperforms previous methods across all datasets, with the most significant improvements observed on SSv2. This improvement is likely due to the alignment of sampled frames with the dataset's characteristics, which allows for the extraction of rich semantic attributes. On K400, the improvement is less pronounced, potentially due to the random sampling of 5 frames per video, which may not capture temporal dynamics as effectively. We also provide self-attention map visualizations in the supplementary material, illustrating how CrossVideoMAE encourages the model to focus on semantically relevant visual regions.

\begin{table}[h!]
\vspace{-5pt}
\centering
\scriptsize
\resizebox{0.8\linewidth}{!}{
\begin{tabular}{lcc}
\toprule
\textbf{Method} & \multicolumn{2}{c}{\textbf{Acc@1 (\%)}} \\
& \textbf{IN-1K}~\cite{russakovsky2015imagenet} & \textbf{SSv2}~\cite{goyal2017something} \\
\midrule
MAE~\cite{he2022masked} / SpatioTemporal MAE~\cite{feichtenhofer2022masked} & 83.60 & 70.0 \\
\rowcolor[gray]{.9}
\textbf{CrossVideoMAE (Ours)} & \textbf{83.62} & \textbf{73.7}\\
\bottomrule
\end{tabular}
}
\vspace{-5pt}
\captionsetup{font=footnotesize}
\caption{Performance on 1N-1K and SSv2 datasets when combining pre-trained MAE and SpatioTemporalMAE with contrastive learning.}
\label{tab:image_encoder}
\vspace{-10pt}
\end{table}

\noindent \textbf{Image Encoder on Action Recognition:} Experiments on IN-1K~\cite{deng2009imagenet} in Tab.~\ref{tab:image_encoder} demonstrate the capabilities of the pre-trained MAE~\cite{he2022masked}. The performance gain in action recognition on the IN-1K dataset is significantly lower than that on the SSv2 dataset. This difference is likely due to the superior accuracy of temporal information in videos. The integration of spatial representation and motion trajectory in videos provides an advantage in motion analysis for action recognition tasks.

\subsection{Analysis and Ablation Studies}

We conduct ablation studies to validate the effectiveness of CrossVideoMAE. Starting with the pre-trained MAE~\cite{he2022masked} ViT-B/16, we pre-train the video encoder using CrossVideoMAE, then fine-tune it under supervised conditions for all SSv2 experiments. (\textit{See supplementary for additional results})

\noindent\textbf{Number of corresponding sampled frames $(n)$:} 
Tab.~\ref{tab:sampled} examines the impact of the image branch by varying the number of sampled frames. When sampling more than one frame, we compute the mean feature embedding of visible tokens across frames for frame-level cross-modal contrastive learning. CrossVideoMAE effectively captures cross-modal frame-level correspondences with just a single sampled frame, enhancing performance. However, with more than two frames, the added information from the image modality may become redundant.

\begin{table}[h!]
\vspace{-5pt}
\centering
\scriptsize
\resizebox{0.8\linewidth}{!}{%
\begin{tabular}{c>{\columncolor[gray]{0.9}}ccccc}
\toprule
\textbf{No. of sampled} & \cellcolor[gray]{0.9} & \multirow{2}{*}{2} & \multirow{2}{*}{3} & \multirow{2}{*}{4} & \multirow{2}{*}{5} \\
\textbf{frame images (n)} & \multirow{-2}{*}{\textbf{1}} \\
\midrule
\textbf{Acc@1 (\%)} & \textbf{73.7} & \textbf{73.7} & 73.5 & 73.4 & 73.1 \\
\bottomrule
\end{tabular}%
}
\vspace{-5pt}
\captionsetup{font=footnotesize}
\caption{Action classification results on SSv2 show that CrossVideoMAE with a single sampled frame (n=1) performs as well or better than using multiple frames. We use n=1 in all experiments.}
\label{tab:sampled}
\vspace{-10pt}
\end{table}

\noindent\textbf{Data Augmentations:}
While self-supervised MAEs~\cite{he2022masked,tong2022videomae,feichtenhofer2022masked} generally use multi-scale cropping alone for pre-training, we explored the effect of additional augmentations, as shown in Tab.~\ref{tab:augment}. We tested random augmentation (resizing, cropping, horizontal flipping), random erasing, mixup, and cutmix. Since masked patches are easier to reconstruct, these augmentations were essential for further performance gains.

\begin{table}[h!]
\vspace{-5pt}
\centering
\scriptsize
\resizebox{\linewidth}{!}{%
\begin{tabular}{cccccc}
\toprule
\textbf{Aug}~\cite{cubuk2020randaugment} & \textbf{Era}~\cite{zhong2020random} & \textbf{MixUp}~\cite{zhang2017mixup} & \textbf{CutMix}~\cite{yun2019cutmix} & \multicolumn{2}{c}{\textbf{Accuracy (\%)}} \\
 & & & & \textbf{Acc@1} & \textbf{Acc@5} \\
\midrule
\ding{55} & $\checkmark$ & $\checkmark$ & $\checkmark$ & 73.72 & 92.67 \\
$\checkmark$ & \ding{55} & $\checkmark$ & $\checkmark$ & 73.51 & \textbf{92.94} \\
$\checkmark$ & $\checkmark$ & \ding{55} & $\checkmark$ & 73.34 & 92.85 \\
$\checkmark$ & $\checkmark$ & $\checkmark$ & \ding{55} & 73.46 & 92.79 \\
\rowcolor[gray]{0.9}
$\checkmark$ & $\checkmark$ & $\checkmark$ & $\checkmark$ & \textbf{73.69} & 92.86 \\
\bottomrule
\end{tabular}%
}
\vspace{-5pt}
\captionsetup{font=footnotesize}
\caption{Performance comparison of various data augmentation techniques on the SSv2 dataset for RandomAugment (Aug), Random Erasing (Era), MixUp, and CutMix respectively.}
\label{tab:augment}
\vspace{-10pt}
\end{table}

\noindent\textbf{Masking Ratios:}
As shown in Tab.~\ref{tab:ratios}, CrossVideoMAE achieves optimal performance with masking ratios of 95\% and 90\%, 5\% and 15\% higher than those in SpatioTemporalMAE~\cite{feichtenhofer2022masked} and MAE~\cite{he2022masked}. These higher ratios enhance representation learning by leveraging the added variation from aggressive masking. In contrast, lowering the masking ratio increases visible tokens, limiting the network’s ability to capitalize on distinctions introduced by high masking, reducing its capacity to capture semantic features from sampled frames.

\begin{table}[h!]
\centering
\resizebox{0.9\linewidth}{!}{
\begin{tabular}[t]{cccc}
\toprule
\multicolumn{2}{c}{\textbf{Mask Ratios}} & \multicolumn{2}{c}{\textbf{Acc@1 (\%)}} \\
\textbf{Image Branch} & \textbf{Video Branch} & \textbf{IN-1K}~\cite{russakovsky2015imagenet} & \textbf{SSv2}~\cite{goyal2017something} \\
\midrule
75\% & 75\% & 83.2 & 72.9 \\
75\% & 90\% & 83.5 & 73.2 \\
75\% & 95\% & 83.4 & 73.4 \\
90\% & 75\% & 83.3 & 72.6 \\
90\% & 90\% & 83.5 & 73.5 \\
\rowcolor[gray]{.9}
\textbf{90\%} & \textbf{95\%} & \textbf{83.6} & \textbf{73.7} \\
\bottomrule
\end{tabular}%
}
\vspace{-5pt}
\caption{Impact of different masking ratios on the image and video branches for action classification accuracy.}
\label{tab:ratios}
\vspace{-10pt}
\end{table}

\noindent\textbf{Impact of Joint Learning Objective:}
As shown in Tab.~\ref{tab:joint} and Section~\ref{sec:method}, our joint feature embedding strategy enhances the model's ability to capture correlations across frame sequences and full videos. By combining intra-modal and cross-modal contrastive learning at both frame and video levels, the model achieves more transferable representations. Ablation studies on SSv2 indicate that removing cross-modal and intra-modal contrastive learning reduces accuracy by 0.7 percentage points (\%p) and 0.5\%p, respectively. Additionally, omitting frame- or video-level objectives results in further accuracy drops of 0.3\%p and 0.4\%p.

\begin{table}[h!]
\vspace{-5pt}
\centering
\scriptsize
\resizebox{\linewidth}{!}{%
\begin{tabular}{ccccc}
\toprule
\multicolumn{2}{c}{\textbf{Different Modal}} & \multicolumn{2}{c}{\textbf{Different Level}} & \textbf{Acc@1 (\%)} \\
\textbf{Intra Modal} & \textbf{Cross Modal} & \textbf{Video Level} & \textbf{Frame Level} & \\
\midrule
\ding{55} & \ding{55} & \checkmark & \checkmark & 70.94 \\
\checkmark & \ding{55} & \checkmark & \checkmark & 72.96 \\
\ding{55} & \checkmark & \checkmark & \checkmark & 72.87 \\
\checkmark & \checkmark & \ding{55} & \ding{55} &72.65 \\
\checkmark & \checkmark & \ding{55} & \checkmark & 73.28 \\
\checkmark & \checkmark & \checkmark & \ding{55} & 73.39 \\
\rowcolor[gray]{0.9}
\checkmark & \checkmark & \checkmark &\checkmark & \textbf{73.70} \\
\bottomrule
\end{tabular}%
}
\vspace{-5pt}
\captionsetup{font=footnotesize}
\caption{Effect of different modalities and levels on classification accuracy using a joint learning objective}
\label{tab:joint}
\vspace{-10pt}
\end{table}

\noindent\textbf{Transfer Learning:}
Tab.~\ref{tab:transfer} showcases the transfer learning effectiveness of our CrossVideoMAE pre-trained model across different datasets for action classification. When fine-tuned on SSv2, our K400-pretrained model achieves a state-of-the-art 73.5\% Acc@1. Similarly, with SSv2 pre-training, it attains 83.0\% Acc@1 on K400, outperforming other MAEs.

\begin{table}[h!]
\centering
\scriptsize
\resizebox{0.8\linewidth}{!}{%
\begin{tabular}[t]{lcccc}
\toprule
\textbf{Pre-train Set} & \textbf{\# Pre-train Data} & \textbf{Fine-tune Set} & \textbf{Acc@1 (\%)} \\
\midrule
K400 & 240k & SSv2 & 73.5 \\
SSv2 & 169k & K400 & 83.0 \\
\bottomrule
\end{tabular}%
}
\vspace{-5pt}
\captionsetup{font=footnotesize}
\caption{ Performance comparison of domain adaptation/transfer learning on different datasets using various pre-training methods.}
\label{tab:transfer}
\vspace{-10pt}
\end{table}

\section{Conclusion}
\label{sec:conclusion}
In this paper, we introduce CrossVideoMAE, an effective end-to-end SSL framework for cross-modal contrastive spatiotemporal and semantic representation learning. By leveraging relationships between videos and sampled frames, our method captures rich spatiotemporal and semantic representations. CrossVideoMAE employs both intra-modal and cross-modal contrastive learning, contrasting features at video and frame levels. Experimental results demonstrate that CrossVideoMAE outperforms previous SOTA methods.

{
    \small
    \bibliographystyle{ieeenat_fullname}
    \bibliography{main}
}

\clearpage
\setcounter{page}{1}
\maketitlesupplementary

\setcounter{section}{0}
\setcounter{figure}{0}   
\setcounter{table}{0}   
\renewcommand{\thesection}{\Alph{section}}

\noindent We organize the Supplementary Materials as follows:

\begingroup
\begin{itemize}
\item \textbf{The overall architecture of our proposed method \S~\ref{sec:overall}}
\item \textbf{The implementation details \S~\ref{sec:implement}.}
\item \textbf{Additional experimental results and analysis \S~\ref{sec:additional} }
\end{itemize}
\endgroup

\section{Overall Architecture of CrossVideoMAE}
\label{sec:overall}

\begin{table*}[!h]
\centering
\resizebox{\linewidth}{!}{
\begin{tabular}[t]{lcccc}
\toprule
\textbf{Stage} & \multicolumn{2}{c}{\textbf{ViT-Base/16 Configuration}} & \multicolumn{2}{c}{\textbf{Output Sizes}} \\ 
& \textbf{Image Branch} & \textbf{Video Branch} & \textbf{Image Branch} & \textbf{Video Branch} \\
& Pre-trained MAE~\cite{he2022masked} & SpatioTemporalMAE~\cite{feichtenhofer2022masked} & & \\
\midrule
Input Image/Video & \ding{55} & 
\begin{tabular}[c]{c} 
Stride $\textcolor{blue}{4} \times \textcolor{teal}{1} \times \textcolor{teal}{1}$ on K400 \\ 
Stride $\textcolor{blue}{2} \times \textcolor{teal}{1} \times \textcolor{teal}{1}$ on SSv2 
\end{tabular} 
& $\textcolor{red}{3} \times \textcolor{teal}{224} \times \textcolor{teal}{224}$ 
& $\textcolor{red}{3} \times \textcolor{blue}{16} \times \textcolor{teal}{224} \times \textcolor{teal}{224}$ \\ 

Patch Embedding & $\textcolor{red}{3} \times \textcolor{teal}{16} \times \textcolor{teal}{16}$, Embedding Dim. $\textcolor{red}{768}$ & 
\begin{tabular}[c]{c} 
$\textcolor{blue}{2} \times \textcolor{red}{3} \times \textcolor{teal}{16} \times \textcolor{teal}{16}$, Embedding Dim. $\textcolor{red}{768}$ \\ 
Stride $\textcolor{blue}{2} \times \textcolor{teal}{16} \times \textcolor{teal}{16}$ 
\end{tabular} 
& $\textcolor{red}{768} \times \textcolor{teal}{14} \times \textcolor{teal}{14}$ 
& $\textcolor{red}{768} \times \textcolor{blue}{8} \times \textcolor{teal}{14} \times \textcolor{teal}{14}$ \\ 

Mask & 
\begin{tabular}[c]{c} 
Random Mask \\ Mask Ratio = $\textcolor{orange}{\rho}$ 
\end{tabular} 
& 
\begin{tabular}[c]{c} 
Random Mask \\ Mask Ratio = $\textcolor{orange}{\rho}$ 
\end{tabular} 
& $\textcolor{red}{768} \times [\textcolor{teal}{14} \times \textcolor{teal}{14} \times (1 - \textcolor{orange}{\rho})]$ 
& $\textcolor{red}{768} \times \textcolor{blue}{8} \times [\textcolor{teal}{14} \times \textcolor{teal}{14} \times (1 - \textcolor{orange}{\rho})]$ \\ 

Encoder & 
$\left[\begin{array}{c} 
\mathrm{MHA}(\textcolor{red}{768}) \\ 
\mathrm{MLP}(\textcolor{red}{3072}) 
\end{array} \right] \times 12$ 
& 
$\left[\begin{array}{c} 
\mathrm{MHA}(\textcolor{red}{768}) \\ 
\mathrm{MLP}(\textcolor{red}{3072}) 
\end{array} \right] \times 12$ 
& $\textcolor{red}{768} \times [\textcolor{teal}{14} \times \textcolor{teal}{14} \times (1 - \textcolor{orange}{\rho})]$ 
& $\textcolor{red}{768} \times \textcolor{blue}{8} \times [\textcolor{teal}{14} \times \textcolor{teal}{14} \times (1 - \textcolor{orange}{\rho})]$ \\ 

Encoder Embedding & 
\begin{tabular}[c]{c} 
$\mathrm{MLP}(\textcolor{red}{384})$ \\ 
\textit{concat learnable tokens} 
\end{tabular} 
& 
\begin{tabular}[c]{c} 
$\mathrm{MLP}(\textcolor{red}{384})$ \\ 
\textit{concat learnable tokens} 
\end{tabular} 
& $\textcolor{red}{384} \times \textcolor{teal}{14} \times \textcolor{teal}{14}$ 
& $\textcolor{red}{384} \times \textcolor{blue}{8} \times \textcolor{teal}{14} \times \textcolor{teal}{14}$ \\ 

Decoder & \ding{55} & 
$\left[\begin{array}{c} 
\mathrm{MHA}(\textcolor{red}{384}) \\ 
\mathrm{MLP}(\textcolor{red}{1536}) 
\end{array} \right] \times 4$ 
& \ding{55} 
& $\textcolor{red}{384} \times \textcolor{blue}{8} \times [\textcolor{teal}{14} \times \textcolor{teal}{14} \times (1 - \textcolor{orange}{\rho})]$ \\ 

Decoder Embedding & \ding{55} & 
$\mathrm{MLP}(\textcolor{red}{1536})$ 
& \ding{55} 
& $\textcolor{red}{1536} \times \textcolor{blue}{8} \times \textcolor{teal}{14} \times \textcolor{teal}{14}$ \\ 

Reshape & \ding{55} & 
from $\textcolor{red}{1536}$ to $\textcolor{red}{3} \times \textcolor{blue}{2} \times \textcolor{teal}{16} \times \textcolor{teal}{16}$ 
& \ding{55} 
& $\textcolor{red}{3} \times \textcolor{blue}{16} \times \textcolor{teal}{224} \times \textcolor{teal}{224}$ \\ 
\bottomrule
\end{tabular}%
}
\caption{\textbf{Encoder and Decoder Architectural Details of CrossVideoMAE.} We take 16-frame vanilla shared, pre-trained MAE ViT-B/16. "MHA" denotes joint space-time self-attention. The output sizes are denoted by $\{\textcolor{red}{C}\times\textcolor{blue}{T}\times\textcolor{teal}{S}\}$ for channel, temporal, and spatial sizes.}
\label{tab:table1}
\end{table*}

\begin{table*}[h!]
\centering
\scriptsize
\resizebox{0.6\linewidth}{!}{
\begin{tabular}[t]{lccc}
\toprule
\textbf{config} & \textbf{Image Branch} & \multicolumn{2}{c}{\textbf{Video Branch}} \\
& \textbf{IN-1K}~\cite{russakovsky2015imagenet} & \textbf{K400}~\cite{kay2017kinetics} & \textbf{SSv2}~\cite{goyal2017something} \\
\midrule
optimizer & AdamW~\cite{loshchilov2018decoupled} & \multicolumn{2}{c}{AdamW~\cite{loshchilov2018decoupled}} \\
base learning rate & 1.5e-4 & \multicolumn{2}{c}{1.5e-4} \\
weight decay & 0.05 & \multicolumn{2}{c}{0.05} \\
optimizer momentum & $\beta_1, \beta_2=0.9, 0.95$~\cite{chen2020generative} & \multicolumn{2}{c}{$\beta_1, \beta_2=0.9, 0.95$~\cite{chen2020generative}} \\
learning rate schedule & cosine decay~\cite{loshchilov2016sgdr} & \multicolumn{2}{c}{cosine decay~\cite{loshchilov2016sgdr}} \\
warmup epochs~\cite{goyal2017accurate} & 40 & \multicolumn{2}{c}{40} \\
Augmentations: & & & \\
 ShortSideScale & N/A & \multicolumn{2}{c}{256px} \\
 RandomResizedCrop & & & \\
\qquad size & 224px & \multicolumn{2}{c}{224px} \\
\qquad scale & [0.08, 1.0] & \multicolumn{2}{c}{[0.08, 1.0]} \\
\qquad ratio & [0.75, 1.33] & \multicolumn{2}{c}{[0.75, 1.33]} \\
\qquad interpolation & Bicubic & \multicolumn{2}{c}{Bilinear} \\
 RandomHorizontalFlip & $\rho$ = 0.5 & $\rho$ = 0.5 & $\rho$ = 0 \\
 Normalize & yes & \multicolumn{2}{c}{yes} \\
\bottomrule
\end{tabular}
}

\caption{Pre-training setting on IN-1K, K400 and SSv2 datasets.}
\label{tab:pre-train}
\end{table*}

\begin{table*}[h!]
\centering
\scriptsize
\resizebox{0.8\linewidth}{!}{%
\begin{tabular}[t]{lcccc}
\toprule
\textbf{config} & \textbf{Image Branch} & \multicolumn{3}{c}{\textbf{Video Branch}} \\
& \textbf{IN-1K}~\cite{russakovsky2015imagenet} & \textbf{K400}~\cite{kay2017kinetics} & \textbf{SSv2}~\cite{goyal2017something} & \textbf{UCF101~\cite{soomro2012ucf101} + HMDB51~\cite{kuehne2011hmdb}} \\
\midrule
optimizer & AdamW & \multicolumn{3}{c}{AdamW} \\
base learning rate & 1e-3 & 5e-4 & 1e-3 & 1.5e-4 \\
weight decay & 0.05 & \multicolumn{3}{c}{0.05} \\
optimizer momentum & $\beta_1=0.9, \beta_2=0.999$ & \multicolumn{3}{c}{$\beta_1=0.9, \beta_2=0.999$} \\
learning rate schedule & cosine decay & \multicolumn{3}{c}{cosine decay} \\
warmup epochs & 5 & \multicolumn{3}{c}{5} \\
Augmentations: & & & \\
 ShortSideScale & N/A & \multicolumn{3}{c}{256px} \\
 RandomResizedCrop & & & \\
\qquad size & 224px & \multicolumn{3}{c}{224px} \\
\qquad scale & [0.08, 1.0] & \multicolumn{3}{c}{[0.08, 1.0]} \\
\qquad ratio & [0.75, 1.33] & \multicolumn{3}{c}{[0.75, 1.33]} \\
\qquad interpolation & Bicubic & \multicolumn{3}{c}{Bilinear} \\
 Repeated Augmentation~\cite{hoffer2020augment} & N/A & \multicolumn{3}{c}{2} \\
 RandomHorizontalFlip & $\rho$ = 0.5 & $\rho$ = 0.5 & $\rho$ = 0 & $\rho$ = 0.5 \\
 RandAugment~\cite{cubuk2020randaugment} & & \\
\qquad magnitude & 9 & \multicolumn{3}{c}{9} \\
\qquad num\_layers & 0.5 & \multicolumn{3}{c}{0.5} \\
 RandomErasing & $\rho$ = 0.25 & $\rho$ = 0 & $\rho$ = 0.25 & $\rho$ = 0.5 \\
 Normalize & yes & \multicolumn{3}{c}{yes} \\
 label smoothing~\cite{szegedy2016rethinking} & 0.1 & \multicolumn{3}{c}{0.1} \\
 mixup~\cite{zhang2017mixup} & 0.8 & \multicolumn{3}{c}{0.8} \\
 cutmix~\cite{yun2019cutmix} & 1.0 & \multicolumn{3}{c}{1.0} \\
 drop path & 0.1 & \multicolumn{3}{c}{0.1} \\
 dropout & 0.1 & \multicolumn{3}{c}{0.1} \\
layer-wise lr decay~\cite{bao2022beit,clark2020electra} & 0.75 & \multicolumn{3}{c}{0.75} \\ 
\bottomrule
\end{tabular}%
}
\caption{End-to-end fine-tuning setting on IN-1K, K400 and SSv2 datasets.}
\label{tab:fine-tune}
\end{table*}

\subsection{Video Branch and Image Branch} 

\subsubsection{Video Branch}

Given a video, we first perform data augmentation to obtain an augmented version of the video. 

\noindent\textbf{Tokenizer:} Given an input video \( u \) of size \( \textcolor{blue}{T} \times \textcolor{red}{C} \times \textcolor{teal}{H} \times \textcolor{teal}{W} \), where \( \textcolor{blue}{T} \) represents the temporal sequence length (frames), \( \textcolor{red}{C} \) is the number of channels, and \( \textcolor{teal}{H}, \textcolor{teal}{W} \) are the spatial dimensions (height and width), we first process it using a patch embedding operation. This involves passing \( u \) through a 3D convolutional layer with a kernel of size \( K = (\textcolor{blue}{t}, \textcolor{red}{C}, \textcolor{teal}{h}, \textcolor{teal}{w}) \), where \( \textcolor{blue}{t} \), \( \textcolor{teal}{h} \), and \( \textcolor{teal}{w} \) define the temporal stride, height, and width dimensions of the kernel, respectively. The convolution uses a stride \( S = (\textcolor{blue}{t}, \textcolor{teal}{h}, \textcolor{teal}{w}) \) and outputs \( \textcolor{red}{D} \) channels. This operation embedding the input video into \( N_u = \textcolor{blue}{\frac{T}{t}} \times \textcolor{teal}{\frac{H}{h}} \times \textcolor{teal}{\frac{W}{w}} \) tokens, each represented as a vector of dimension \( D \).

\noindent\textbf{Positional Encoding:} Positional information is then added to the tokens \( N_u \) to retain their spatial and temporal context.

\noindent\textbf{Masking:} Randomly mask \( M_u \) tokens out of the total \( N_u \) tokens.

\noindent\textbf{Encoder:} Next, we generate feature embedding \(f_{\theta_u}(\cdot)\) of visible tokens by passing \(N_u - M_u\) visible tokens with positional information through the transformer ViTEncoder.

\noindent\textbf{Decoder:} The feature embedding of the visible tokens is concatenated with a set of fixed, learnable feature embeddings of the masked tokens \( M_u \) to generate the combined embeddings. Positional encodings are then added to both the visible and masked token embeddings. This combined representation is passed through a lightweight transformer-based ViTDecoder, which is trained using the Mean Squared Error (MSE) loss. The loss is computed between the reconstructed tokens of the video and its augmented counterpart, denoted as \( \tilde{u}_i \) and \( \tilde{u}_i^t \), ensuring accurate reconstruction of the input tokens.

\subsubsection{Frame Image Branch}

Similarly, for the image branch, a set of random frames is manually sampled from the video to generate corresponding sampled frame images.

\noindent\textbf{Tokenizer:} Given an input sampled frame \( f \) of size \(\textcolor{red}{C} \times \textcolor{teal}{H} \times \textcolor{teal}{W}\), where \(\textcolor{teal}{H}\) and \(\textcolor{teal}{W}\) represent the spatial dimensions and \(\textcolor{red}{C}\) denotes the number of channels, we first pass \( f \) through a Patch Embedding layer. This layer is implemented as a 3D convolution with a kernel size of \( K_f = (\textcolor{red}{C}, \textcolor{teal}{h}, \textcolor{teal}{w}) \), producing \(\textcolor{red}{D}\) output channels. This operation embeds \( f \) into \( N_f = \textcolor{teal}{\frac{H}{h}} \times \textcolor{teal}{\frac{W}{w}} \) tokens, each with a dimension of \( D \).

\noindent\textbf{Positional Encoding:} Positional information is then added to the tokens \(N_f\) to retain their spatial context.

\noindent\textbf{Masking:} Randomly mask \( M_f \) tokens out of the total \( N_f \) tokens.

\noindent\textbf{Encoder:} Next, we generate feature embedding \(f_{\theta_f}(\cdot)\) by passing \(N_f - M_f\) visible tokens with positional information through the pre-trained MAE~\cite{he2022masked} transformer ViTEncoder.

\subsection{Architecture Details}
\label{sec:detailed}

Tab.~\ref{tab:table1} details the architecture of the encoder and decoder of our CrossVideoMAE. Specifically, we take the 16-frame vanilla shared, pre-trained ViT-B/16 for all experiments. We use an asymmetric encoder-decoder architecture for self-supervised cross-modal video pre-training and discard the decoder during the fine-tuning phase. We adopt the joint space-time attention~\cite{arnab2021vivit,liu2022video} to capture the rich spatiotemporal representations and semantic attributes in the visible tokens. 

Given a video, we first extract \textcolor{blue}{16} frames (\textcolor{red}{3} $\times$ \textcolor{blue}{16} $\times$ \textcolor{teal}{224} $\times$ \textcolor{teal}{224}). These frames are extracted uniformly at regular intervals for both datasets, as outlined in previous work~\cite{feichtenhofer2022masked}. We use a temporal stride of 4 and 2 for the K400 and SSv2 datasets, respectively. Next, we process this \textcolor {blue}{16} frames through Patch Embedding, which is essentially a convolution layer with a kernel size of \textcolor{blue}{2} $\times$ \textcolor{red}{3} $\times$ \textcolor{teal}{16} $\times$ \textcolor{teal}{16}, the stride of \textcolor{blue}{2}$\times$\textcolor{teal}{16}$\times$\textcolor{teal}{16}, and output embedding dimension of \textcolor{red}{768}.This process results in a total of \textcolor{purple}{1568} tokens, and each token is represented by a \textcolor{red}{768} dimensional vector. A standard positional encoding vector is added to the embedded patches. Next, we mask $M_u = \textcolor{orange}{\rho_u}\times\textcolor{purple}{1568}$ number of tokens and proceed $N_u - M_u = (1 - \textcolor{orange}{\rho_u})\times\textcolor{purple}{1568}$ as the visible tokens. $\rho_u$ denotes the masking ratio applied to the video branch. These visible tokens are then processed through the shared MAE ViT video encoder that comprises 12 cascaded multi-head self-attention blocks (MHA blocks). The shared MAE ViT video encoder outputs are then concatenated with a fixed learnable representation for masked tokens, resulting in the \textcolor{purple}{1568} token representations. This \textcolor{purple}{1568} representations are then processed through an encoder embedding which brings down their embedding dimension from \textcolor{red}{768} to \textcolor{red}{384} by an MLP layer. These embedded representations are then processed through the shared MAE ViT-decoder which consists of 4 MHA blocks followed by an MLP layer to bring the embedding dimension from \textcolor{red}{384} to \textcolor{red}{1536} to compute the MSE loss, and the total number of pixels in a cube which is given by $\textcolor{red}{2}\times\textcolor{blue}{3} \times\textcolor{teal}{16}\times\textcolor{teal}{16} = \textcolor{red}{1536}$. This is finally reshaped back to the original space and used to compute the reconstruction loss. 

Given a sampled frame (\textcolor{red}{3} $\times$ \textcolor{teal}{224} $\times$ \textcolor{teal}{224}), we first process this through patch embedding, which is essentially a convolution layer with a kernel size of \textcolor{teal}{16}$\times$\textcolor{teal}{16}, and output embedding dimension of \textcolor{red}{768}. A standard positional encoding vector is added to the embedded patches and fed into the encoder. This process results in a total of \textcolor{purple}{196} tokens, and each token is represented by a \textcolor{red}{768} dimensional vector. Next, we mask $M_f = \textcolor{orange}{\rho_f}\times\textcolor{purple}{196}$ number of tokens and proceed $N_f - M_f = (1 - \textcolor{orange}{\rho_f})\times\textcolor{purple}{196}$ as the visible tokens. $\rho_f$ denotes the masking ratio applied to the frame image branch. These visible tokens are then processed through the pre-rained MAE~\cite{he2022masked} ViT image encoder that comprises 12 cascaded multi-head self-attention blocks (MHA blocks). These visible tokens are then processed through an encoder embedding which brings down their embedding dimension from \textcolor{red}{768} to \textcolor{red}{384} by an MLP layer. This pre-trained MAE~\cite{he2022masked} ViT image encoder learned visible tokens: $N_f - M_f$ representations are then processed through an encoder embedding which brings down their embedding dimension from \textcolor{red}{768} to \textcolor{red}{384} by an MLP layer.

These encoder-embedded features facilitate spatiotemporal-spatial feature embedding correspondence by maximizing mutual information between video, augmented video, and sampled frames. Visible tokens in the feature-invariant space are processed in a self-supervised fashion, promoting invariance to augmentations in the video domain. Furthermore, this process distills well-learned knowledge from sampled frames to videos through intra-modal, cross-modal, frame-level, and video-level contrastive learning. This approach enables the model to effectively capture visual concepts, ensure view invariance, and extract semantic attributes analogous to human perception.

\section{Implementation Details}
\label{sec:implement}

We followed the pre-training configurations outlined in previous works, such as MAE ~\cite{he2022masked} and SpatioTemporalMAE ~\cite{feichtenhofer2022masked}. 

\subsection{Datasets}

We evaluated our method on four video datasets commonly used for action recognition: 
Kinetics-400 (K400)~\cite{kay2017kinetics} Something-Something V2 (SSv2)~\cite{goyal2017something}, UCF101~\cite{soomro2012ucf101}, and HMDB51~\cite{kuehne2011hmdb}. 

K400: contains video clips from YouTube, around 240k training videos, and 20k validation videos of 10s from 400 action classes.

SSv2: is also a large-scale video dataset, having around 169k videos for training and 20k videos for validation of 4s, categorized into 174 motion-centric action classes. We conducted ablation studies on the SSv2 dataset and reported results on both K400 and SSv2 datasets.

UCF101: is a relatively small dataset, consisting of $\sim$9.5K training videos and $\sim$3.5K validation videos.

HMDB51: is also a small video dataset that contains around 3.5K/1.5K train/val videos. On UCF101 and HMDB51, we follow the commonly used protocols and evaluate our method across all 3 train/val splits. 

ImageNet-1K (IN-1K)~\cite{russakovsky2015imagenet} We use the ILSVRC 2012 challenge subset, which includes 1.28M training and 50K validation images spanning 1000 classes.

\begin{table}[!h]
\centering
\begin{tabular}[t]{cc}
\toprule
\textbf{Config} & \textbf{SSv2} \\ 
\midrule
optimizer & SGD \\ 
base learning rate & 0.1 \\ 
weight decay & 0 \\ 
optimizer momentum & 0.9 \\ 
learning rate schedule & cosine decay \\
warmup epochs & 10 \\ 
training epochs & 100 \\ 
augmentation & MultiScaleCrop \\ 
\bottomrule
\end{tabular}%

\caption{Linear probing setting.}
\label{tab:linear}
\end{table}

\begin{table*}[!t]
\centering
\scriptsize
\resizebox{0.8\linewidth}{!}{%
\begin{tabular}[t]{lcccccccc}
\toprule
\multirow{2}{*}{\textbf{Method}} & \multirow{2}{*}{\textbf{Modality}} & \multirow{2}{*}{\textbf{Backbone}} & \multirow{2}{*}{\textbf{Extra Data}} & \multicolumn{4}{c}{\textbf{Action Linear Classification (Acc@1 (\%))}} \\
& & & &\textbf{UCF101} & \textbf{HMDB51} & \textbf{K400} & \textbf{SSv2} \\

\midrule
MoCo~\cite{he2020momentum} & V & R50 & UCF101 & 65.4 & — & 34.5 & 7.4 \\
CoCLR-RGB~\cite{han2020self} & V & R(2+1)D & UCF101 & 74.5 & 46.1 & — & — \\
CVRL~\cite{qian2021spatiotemporal} & V & SlowOnly-R50 & K400 & 89.8 & 58.3 & 66.1 & — \\
$\rho$BYOL~\cite{feichtenhofer2021large} & V & SlowOnly-R50 & K400 & 90.1 & 61.1 & 68.3 & 24.5 \\
VideoMoCo~\cite{pan2021videomoco} & V & R(2+1)D & K400 66.3 & — & 31.0 & 19.5 \\
CORP\textsubscript{f}~\cite{hu2021contrast}& V & SlowOnly-R50 & K400 & 90.2 & 58.7 & 66.6 & — \\
Vi$^2$CLR~\cite{diba2021vi2clr} & V & S3D & K400 & 75.4 & 47.3 & 63.4 & — \\
GDT~\cite{patrick2021multi}& V + A & R(2+1)D & IG65M & 75.7 & — & 38.6 & 11.9 \\
TimeSformer~\cite{bertasius2021space} & V & ViT-B & IN-21K & — & — & 14.0 & — \\
SVT~\cite{ranasinghe2022self} & V & ViT-B & IN-21K+K400 & 90.8 & 57.8 & \textcolor{blue}{68.1} & 18.3 \\
ViMPAC~\cite{tan2021vimpac} & V + I & ViT-L & HowTo100M+DALLE & — & — \\
VideoMAE~\cite{tong2022videomae} & V & ViT-B & K400 & 84.6 & 60.5 & 61.2 & 23.1 \\
MME~\cite{sun2023masked} & V & ViT-B & K400 & — & — & — & \textcolor{blue}{29.2} \\
MVD-B~\cite{wang2023masked} & V + I & Teacher-B & IN-1K + K400 & \textcolor{blue}{97.0} & \textcolor{blue}{76.4} & — & — \\
\midrule
\rowcolor[gray]{0.9}
\textbf{CrossVideoMAE} & V + I & ViT-B & IN-1K + K400 & \textcolor{red}{97.6} & \textcolor{red}{76.9} & \textcolor{red}{68.7}& \textcolor{red}{31.2}\\
\bottomrule
\end{tabular}%
}
\caption{Comparison with state-of-the-art methods on UCF101, HMDB51, K400 and SSv2 for linear probing. ‘A’ is audio, and ‘I’ is image. The best and second best results are marked by \textcolor{red}{red} and \textcolor{blue}{blue} colours, respectively.}
\label{tab:action_linear}
\end{table*}

\begin{table*}[!h]

\centering
\scriptsize
\resizebox{0.9\linewidth}{!}{%
\begin{tabular}[t]{lccccccccc}
\toprule
\multirow{2}{*}{\textbf{Method}} & \multirow{2}{*}{\textbf{Backbone}} & \textbf{Extra pre-training} & \multirow{2}{*}{\textbf{Extra labels}} & \multirow{2}{*}{\textbf{Frames}} & \textbf{GFLOPs (G)} & \textbf{Param} & \textbf{Acc@1} & \textbf{Acc@5} \\
& & \textbf{dataset} & & & \textbf{FLOPs}$\times$\textbf{Clips}$\times$\textbf{Crops} & \textbf{(M)} & \textbf{(\%)} & \textbf{(\%)} \\
\midrule\midrule
\multicolumn{7}{l}{\textbf{\textit{Category: Supervised Pre-training}}} \\
\midrule
TSM\textsubscript{$two$ $stream$}~\cite{lin2019tsm} & ResNet50\textsubscript{$\times$2} & \multirow{4}{*}{IN-1K} & \checkmark & 16+16 & 130$\times$2$\times$3 & 49 & 66.0 & 90.5 \\
TEINet\textsubscript{$En$}~\cite{liu2020teinet} & ResNet50\textsubscript{$\times$2} & & \checkmark & 8+16 & 99$\times$10$\times$3 & 50 & 66.6 & N/A \\
TANet\textsubscript{$En$}/TAM~\cite{liu2021tam}& ResNet50\textsubscript{$\times$2} & & \checkmark & 8+16 & 99$\times$2$\times$3 & 51 & 66.0 & 90.1 \\
TDN\textsubscript{$En$}~\cite{wang2021tdn} & ResNet101\textsubscript{$\times$2} & & \checkmark & 8+16 & 198$\times$1$\times$3 & 88 & 69.6 & 92.2 \\
\midrule
SlowFast~\cite{feichtenhofer2019slowfast} & ResNet101 & \multirow{2}{*}{K-400} & \checkmark & 8+32 & 106$\times$1$\times$3 & 53 & 63.1 & 87.6 \\
MViTv1~\cite{fan2021multiscale} & MViTv1-B & & \checkmark & 64 & 455$\times$1$\times$3 & 37 & 67.7 & 90.9 \\
\midrule
TimeSformer~\cite{bertasius2021space} & ViT-B & \multirow{2}{*}{IN-21K} & \checkmark & 8 & 196$\times$1$\times$3 & 121 & 59.5 & N/A \\
TimeSformer~\cite{bertasius2021space} & ViT-L & & \checkmark & 64 & 5549$\times$1$\times$3 & 430 & 62.4 & N/A \\
\midrule
ViViT FE~\cite{arnab2021vivit} & ViT-L & IN-21K+K400 & \checkmark & 32 & 995$\times$4$\times$3 & N/A & 65.9 & 89.9 \\
TAdaConvNeXt-T~\cite{huang2021tada} & ConvNeXt-T & IN-1K & \checkmark & 32 & 94$\times$3$\times$2 & 38 & 67.1 & 90.4 \\
\midrule
Motionformer~\cite{patrick2021keeping} & ViT-B & \multirow{3}{*}{IN-21K+K400} & \checkmark & 16 & 370$\times$1$\times$3 & 109 & 66.5 & 90.1 \\
Motionformer~\cite{patrick2021keeping} & ViT-L & & \checkmark & 32 & 1185$\times$1$\times$3 & 382 & 68.1 & 91.2 \\
Video Swin~\cite{liu2020teinet} & Swin-B & & \checkmark & 32 & 321$\times$1$\times$3 & 88 & 69.6 & 92.7 \\
\midrule\midrule
\multicolumn{7}{l}{\textbf{\textit{Category: Self-Supervised Pre-training}}} \\
\midrule
\hdashline
\multicolumn{7}{l}{\textbf{\textit{Pre-trained Epochs: 800}}} \\
\hdashline
\rowcolor[gray]{0.9}
\textbf{CrossVideoMAE (Ours)} & \textbf{ViT-B} & IN-1K & \ding{55} & \textbf{16} & 180$\times$2$\times$3 & 87 (Shared) & \textcolor{red}{73.7} & \textcolor{red}{93.4} \\
\bottomrule
\end{tabular}%
}

\caption{Comparison of our proposed method with supervised SOTA methods on SSv2 dataset. We use ViT-B/16 backbone. Extra labels \ding{55} denotes only unlabeled data used for the pre-training phase. The N/A denotes these numbers as not being available/reported in the paper. The best result is marked by \textcolor{red}{red} colour.}
\label{tab:supervised_ssv2}
\end{table*}

We conduct the experiments with the pre-trained models adopted from open-source repositories (
MAE ~\cite{he2022masked} and SpatioTemporalMAE ~\cite{feichtenhofer2022masked}) and fine-tuning on the K400, SSv2, UCF101, HMDB51, and IN-1K datasets.

\subsection{Pre-training} 

The default settings for pre-training and end-to-end finetuning on IN-1K, K400, and SSv2 datasets are shown in Tab.~\ref{tab:pre-train} and Tab.~\ref{tab:fine-tune}. We use the pre-trained model on the Kinetics-400 [1600 epochs] and then transfer it to the UCF101 and HMDB51. The default
settings of fine-tuning for 100 epochs and 50 epochs, respectively, are shown in Tab.~\ref{tab:fine-tune}.

Tab.~\ref{tab:pre-train} details the pre-training setting on IN-1K, K400, and SSv2 datasets. In addition, we linearly scale the base learning rate w.r.t the overall batch size, $\textit{\text{lr}} = \textit{\text{base\_learning\_rate}} \times \text{\textit{batchsize} / 256}$~\cite{goyal2017accurate}. We adopt the PyTorch and DeepSpeed frameworks for faster training.

\begin{table*}[h!]

\centering
\scriptsize
\resizebox{0.9\linewidth}{!}{%
\begin{tabular}[t]{lccccccccc}
\toprule
\multirow{2}{*}{\textbf{Method}} & \multirow{2}{*}{\textbf{Backbone}} & \textbf{Extra pre-trainining} & \multirow{2}{*}{\textbf{Extra labels}} & \multirow{2}{*}{\textbf{Frames}} & \textbf{GFLOPs (G)} & \textbf{Param} & \textbf{Acc@1} & \textbf{Acc@5} \\
& & \textbf{dataset} & & & \textbf{FLOPs}$\times$\textbf{Clips}$\times$\textbf{Crops} & \textbf{(M)} & \textbf{(\%)} & \textbf{(\%)} \\
\midrule\midrule
\multicolumn{7}{l}{\textbf{\textit{Category: Supervised Pre-training}}} \\
\midrule
\midrule
NonLocal I3D~\cite{wang2018non} & ResNet101 & \multirow{5}{*}{IN-1K} & \checkmark & 128 & 359$\times$10$\times$3 & 62 & 77.3 & 93.3 \\
TAdaConvNeXt-T~\cite{huang2021tada} & ConvNeXt-T & & \checkmark & 32 & 94$\times$3$\times$4 & 38 & 79.1 & 93.7 \\
TANet/TAM~\cite{liu2021tam} & ResNet152 & & \checkmark & 16 & 242$\times$4$\times$3 & 59 & 79.3 & 94.1 \\
TDN\textsubscript{$En$}~\cite{wang2021tdn} & ResNet101\textsubscript{$\times$2} & & \checkmark & 8+16 & 198$\times$10$\times$3 & 88 & 79.4 & 94.4 \\
Video Swin~\cite{liu2020teinet} & Swin-B & & \checkmark & 32 & 282$\times$4$\times$3 & 88 & 80.6 & 94.6 \\
\midrule
TimeSformer~\cite{bertasius2021space} & ViT-B & \multirow{5}{*}{IN-21K} & \checkmark & 8 & 196$\times$1$\times$3 & 121 & 78.3 & 93.7 \\
TimeSformer~\cite{bertasius2021space} & ViT-L & & \checkmark & 96 & 8353$\times$1$\times$3 & 430 & 80.7 & 94.7 \\
ViViT FE~\cite{arnab2021vivit} & ViT-L & & \checkmark & 128 & 3980$\times$1$\times$3 & N/A & 81.7 & 93.8 \\
Motionformer~\cite{patrick2021keeping} & ViT-B & & \checkmark & 16 & 370$\times$10$\times$3 & 109 & 79.7 & 94.2 \\
Motionformer~\cite{patrick2021keeping} & ViT-L & & \checkmark & 32 & 1185$\times$10$\times$3 & 382 & 80.2 & 94.8 \\
Video Swin~\cite{liu2020teinet} & Swin-L & & \checkmark & 32 & 604$\times$4$\times$3 & 197 & 83.1 & 95.9 \\
\midrule
ViViT FE~\cite{arnab2021vivit} & ViT-L & \multirow{2}{*}{JFT-300M} & \checkmark & 128 & 3980$\times$1$\times$3 & N/A & 83.5 & 94.3 \\
ViViT~\cite{arnab2021vivit} & ViT-H & & \checkmark & 32 & 3981$\times$4$\times$3 & N/A & 84.9 & 95.8 \\
\midrule
ip-CSN~\cite{tran2019video} & ResNet152 & \multirow{3}{*}{\textbf{---}} & \ding{55} & 32 & 109$\times$10$\times$3 & 33 & 77.8 & 92.8 \\
SlowFast~\cite{feichtenhofer2019slowfast} & R101+NL & & \ding{55} & 16+64 & 234$\times$10$\times$3 & 60 & 79.8 & 93.9 \\
MViTv1~\cite{fan2021multiscale} & MViTv1-B & & \ding{55} & 32 & 170$\times$5$\times$1 & 37 & 80.2 & 94.4 \\
\midrule
\multicolumn{7}{l}{\textbf{\textit{Category: Self-Supervised Pre-training}}} \\
\midrule
\multicolumn{7}{l}{\textbf{\textit{Pre-Trained Epochs: 1600}}} \\
\rowcolor[gray]{0.9}
\textbf{CrossVideoMAE (Ours)} & ViT-B & 1N-1K & \ding{55} & 16 & 180$\times$7$\times$3 & 87 (Shared) & \textcolor{red}{83.2} & \textcolor{red}{95.6} \\
\bottomrule
\end{tabular}%
}

\caption{Comparison of our proposed method with supervised SOTA methods on the K400 dataset. We use ViT-B/16 backbone. Extra labels \ding{55} denotes only unlabelled data used for the pre-training phase. The N/A denotes these numbers as not being available/reported in the paper. The best result is marked by \textcolor{red}{red} colour.}
\label{tab:supervised_k400}

\end{table*}

\begin{table*}[!h]
\centering
\scriptsize
\resizebox{0.7\linewidth}{!}{%
\begin{tabular}[t]{lccccc}
\toprule
\textbf{Method} & \textbf{Modality} & \textbf{Backbone} & \textbf{Extra Data} & \multicolumn{2}{c}{\textbf{Video Retrieval (R@1)}} \\
& & & & \textbf{UCF101} & \textbf{HMDB51} \\
\midrule
VCOP~\cite{xu2019self} & V & R(2+1)D & UCF101 & 14.1 & — \\
CoCLR-RGB~\cite{han2020self} & V & S3D-G & K400 & 53.3 & 23.2 \\
Vi$^2$CLR~\cite{diba2021vi2clr} & V & S3D & K400 & 55.4 & 24.6 \\
$\rho$BYOL\textsubscript{$\rho=4$}~\cite{feichtenhofer2021large} & V & SlowOnly-R50 & K400 & 76.8 & 39.6 \\
SVT~\cite{ranasinghe2022self} & V & ViT-B & IN-21K+K400 & \textcolor{blue}{82.9} & \textcolor{blue}{44.4} \\
VideoMAE~\cite{tong2022videomae}& V & ViT-B & K400 & 64.0 & 32.5 \\
\midrule
\rowcolor[gray]{.9}
\textbf{CrossVideoMAE} & V + I & ViT-B & IN-1K + K400 & \textcolor{red}{85.5} & \textcolor{red}{49.7} \\
\bottomrule
\end{tabular}%
}

\caption{Comparison with state-of-the-art methods on UCF101 and HMDB51 forVideo Retrieval. ‘V’ refers to visual, ‘A’ is audio, ‘T’ is text narration, and ‘I’ is the image. The best and second best results are shown in \textcolor{red}{red} and \textcolor{blue}{blue} colours, respectively.}
\label{tab:video_Retrival}
\end{table*}

\subsection{Evaluation}
We evaluate our models under two main methods: End-to-end full fine-tuning and linear evaluation.

\subsubsection{End-to-end full Finetuning} 
Default settings for end-to-end fine-tuning can be found in Tab.~\ref{tab:fine-tune} on IN-1K, K400, SSv2, UCF101, and HMDB51 datasets. Similar to previous work, we use layer-wise learning rate decay~\cite{he2022masked}.

\subsubsection{Linear probing} 
We further evaluate our method under liner probing setting on the UCF101, HMDB51, K400, and SSv2 datasets. We follow SVT~\cite{ranasinghe2022self} to fix the transformer backbone and train a linear layer for 100 epochs. Tab.~\ref{tab:linear} shows the settings that we use for linear evaluation.

\section{Additional Results}
\label{sec:additional}

\subsection{Comparison with State-of-the-Art Methods}
In this section, we provide an extended set of results, evaluating our method on action recognition tasks through linear evaluation and full fine-tuning and comparing it against supervised learning models. We also report comparative results on video retrieval tasks.

\subsubsection{Action Recognition}

\paragraph{Linear Evaluation:}

Table~\ref{tab:action_linear} presents the linear evaluation results for action recognition on the UCF101, HMDB51, K400, and SSv2 datasets. Our model, CrossVideoMAE, consistently outperforms the current state-of-the-art methods across all datasets.

\paragraph{End-to-End Full Fine-Tuning (Supervised Learning Evaluation):}

In Tables~\ref{tab:supervised_ssv2} and~\ref{tab:supervised_k400}, we present a comparison of CrossVideoMAE's performance on the SSv2 and K400 datasets against other state-of-the-art methods that rely on supervised pre-training. Our method demonstrates superior performance in both datasets, highlighting its effectiveness for end-to-end fine-tuning.

\subsubsection{Video Retrieval}

Table~\ref{tab:video_Retrival} showcases the results of video retrieval on the UCF101 and HMDB51 datasets. CrossVideoMAE achieves the highest retrieval accuracy on both datasets, with 85.5\% on UCF101 and 49.7\% on HMDB51, setting a new benchmark for video retrieval performance in these tasks.

\subsection{More analysis and ablation studies}

\subsubsection{Sampled frame selection}

In this study, we investigate the influence of sampled frame selection on the distillation process. We compare the random frame as the sampled frame with either the first or a middle frame, and the result is shown in Tab.~\ref{tab:sampled}. This implies that the random frame is the best, as K400/SSv2 dataset videos are short-range (4-10s) videos.

\begin{table}[h!]
\centering
\scriptsize
\resizebox{0.7\linewidth}{!}{
\begin{tabular}[t]{ccc}
\toprule
\textbf{Sampled frame} & \textbf{Acc@1} & \textbf{Acc@5} \\
\midrule
first frame & 73.0 &92.9 \\
middle frame & 73.3 & 93.1 \\
\rowcolor[gray]{0.9}
\textbf{random frame} & \textbf{73.7} & \textbf{93.4} \\
\bottomrule
\end{tabular}%
}

\caption{\textbf{Sampled frame selection.} We perform an ablation study on SSv2 to select the sampled frame as the first, middle, or random frame}
\label{tab:sampled}
\end{table}

\subsubsection{Masking Types} 
We applied random masking to the image branch and tested frame, tube, and random masking for the video branch (Tab.~\ref{tab:types}). Our results showed that random masking in both branches achieved the best performance. Frame masking, which hides entire tokens in random frames, performed poorly might be due to pixel redundancy across frames. Tube masking~\cite{tong2022videomae}, which masks tokens at the same spatial location over consecutive frames, also underperformed as it might struggle to transfer learned semantics effectively. Random patch masking~\cite{feichtenhofer2022masked} with high ratios (90-95\%) worked well for both images and videos, hence we selected random masking for both modalities.

\begin{table}[h!]
\centering
\resizebox{0.9\linewidth}{!}{
\begin{tabular}[t]{cccc}
\toprule
\multicolumn{2}{c}{\textbf{Masking Types}} & \multicolumn{2}{c}{\textbf{Acc@1 (\%)}} \\
\textbf{Image Branch} & \textbf{Video Branch} & \textbf{IN-1K}~\cite{russakovsky2015imagenet} & \textbf{SSv2}~\cite{goyal2017something} \\
\midrule
Random & Tube & 83.4 & 73.4 \\
Random & Frame & 83.1 & 72.7 \\
\rowcolor[gray]{.9}
\textbf{Random} & \textbf{Random} & \textbf{83.6} & \textbf{73.7} \\
\bottomrule
\end{tabular}%
}

\caption{Performance comparison of various masking strategies on the IN-1K SSv2 dataset using Acc@1, highlighting the impact of different combinations of image and video branches.}
\label{tab:types}

\end{table}

\subsubsection{Decoder Depth}Tab.~\ref{tab:decoder} illustrates the impact of varying decoder depths on action classification accuracy.The results indicate that increasing the number of decoder blocks generally improves accuracy, with four blocks achieving the highest performance. However, using eight blocks slightly decreases top-1 accuracy, suggesting diminishing returns beyond four blocks.

\begin{table}[h!]
\centering
\scriptsize
\resizebox{0.6\linewidth}{!}{%
\begin{tabular}{ccc}
\toprule
\textbf{Blocks} & \multicolumn{2}{c}{\textbf{Accuracy (\%)}} \\
& \textbf{Acc@1 (\%)} & \textbf{Acc@5 (\%)} \\
\midrule
1 & 72.52 & 92.65 \\
2 & 72.79 & 92.87 \\
\rowcolor[gray]{0.9}
\textbf{4} & \textbf{73.70} & \textbf{93.40} \\
8 & 71.63 & 93.35 \\
\bottomrule
\end{tabular}%
}
\caption{Impact of varying decoder depth on action classification accuracy.}
\label{tab:decoder}
\end{table}

\subsubsection{Further analysis of the impact of joint learning objective} We emphasize that addressing both intra-modal and cross-modal contrastive learning in a joint manner contributes to richer representation learning than individual objectives alone. Besides, both video and frame-level contrastive learning capture spatial and spatio-temporal prior representations. 

Intra-modal contrastive learning encourages the model to capture the spatiotemporal correspondence by imposing invariance to augmentations, while cross-modal contrastive learning establishes spatiotemporal-spatial correspondence and fine-grained part semantic attributes. Video-level and frame-level contrastive learning capture spatio-temporal prior and spatial prior representations, respectively.

\begin{table}[h!]
\centering
\resizebox{0.9\linewidth}{!}{
\begin{tabular}{lc}
\toprule
\textbf{Contrastive Learning Technique} & \textbf{Acc@1. Drop (\%)} \\ 
\midrule
Without Intra-Modal Contrastive Learning & 0.5 \\ 
Without Cross-Modal Contrastive Learning & 0.7 \\ 
Without Cross-Modal + Intra-Modal Contrastive Learning & 1.2 \\ 
Without Frame Level Contrastive Learning & 0.3 \\
Without Video Level Contrastive Learning & 0.4 \\ 
Without Video Level + Frame Level Contrastive Learning & 0.7 \\ 
\bottomrule
\end{tabular}%
}

\caption{Effect of the joint learning objective on intra-modal, cross-modal, frame-level, and video-level tasks. Action recognition performance of pre-trained embeddings evaluated on the SSv2 dataset under the default configuration.}
\label{tab:table8}

\end{table}

We empirically test this by conducting ablation studies on the SSv2 dataset, training the model in all possible settings, and evaluating its performance on action recognition. Our findings, as shown in Tab.~\ref{tab:table8}, illustrate that in all learning settings, the proposed joint learning paradigm outperforms the individual objectives. Notably, the combination of both intra-modal and cross-modal, and both video and frame-level learning objectives, obtain an accuracy gain of 0.8\% over the second best approach in SSv2 with the pre-trained SpatioTemporalMAE~\cite{feichtenhofer2022masked} video encoder.

\subsubsection{Effect of corresponding data.} Since SpatioTemporalMAE is pre-trained with a sampled frame image dataset instead of IN-1K, one concern is whether the gains can be attributed to joint training. To that end, we experiment with a pre-training image branch (pre-trained MAE) with IN-1K instead of the sampled frame dataset. To ensure, we use the exact setup for CrossVideoMAE: ensuring the exact same epochs, number of parameter updates, data, learning rates schedule, etc. As shown in Tab.~\ref{tab:table9}, the SSv2 video action recognition performance drops significantly by almost 2.9\% when trained using the IN-1K dataset. This shows that the performance gains with CrossVideoMAE are not merely due to the IN-1K being used for training. This ensures that the gains are indeed from jointly training on the corresponding two modality datasets rather than simply using more data during training.

\begin{table}[h!]
\centering
\resizebox{\linewidth}{!}{
\begin{tabular}{lccc}
\toprule
\textbf{Setting} & \textbf{Data} & \multicolumn{2}{c}{\textbf{Performance (\%)}} \\
& & \textbf{IN-1K}~\cite{russakovsky2015imagenet} & \textbf{SSv2}~\cite{goyal2017something} \\
\midrule
\multirow{2}{*}{\textbf{CrossVideoMAE (Ours)}} & IN-1K + SSv2 & 82.8 & 70.8 \\
& sampled frame dataset + SSv2 & 83.1 & 73.7 \\
\bottomrule
\end{tabular}
}
\caption{Effect of corresponding data}
\label{tab:table9}

\end{table}

\subsubsection{Masking Types} 

\begin{table}[h!]
\centering
\resizebox{0.9\linewidth}{!}{
\begin{tabular}[t]{cccc}
\toprule
\multicolumn{2}{c}{\textbf{Masking Types}} & \multicolumn{2}{c}{\textbf{Acc@1 (\%)}} \\
\textbf{Image Branch} & \textbf{Video Branch} & \textbf{IN-1K}~\cite{russakovsky2015imagenet} & \textbf{SSv2}~\cite{goyal2017something} \\
\midrule
Random & Tube & 83.4 & 73.4 \\
Random & Frame & 83.1 & 72.7 \\
\rowcolor[gray]{.9}
\textbf{Random} & \textbf{Random} & \textbf{83.6} & \textbf{73.7} \\
\bottomrule
\end{tabular}%
}

\caption{Performance comparison of various masking strategies on the IN-1K SSv2 dataset using Acc@1, highlighting the impact of different combinations of image and video branches.}
\label{tab:types}

\end{table}
We applied random masking to the image branch and explored frame, tube, and random masking for the video branch (Tab.~\ref{tab:types}). Our experiments revealed that random masking for both branches yielded the best performance. Frame masking, which masks entire tokens in randomly selected frames, performed worse due to high pixel redundancy across frames. Tube masking~\cite{tong2022videomae}, which masks tokens at the same spatial location across consecutive frames, was also less effective, as it struggled to transfer well-learned semantic information from the sampled frames to full videos. Consequently, we opted for random masking in both branches. Additionally, random patch masking~\cite{feichtenhofer2022masked}, which masks tokens randomly across space and time, performed well with high masking ratios (90\% and 95\%) in both images and videos. Given its simplicity and effectiveness, we chose random masking for both modalities.

\subsection{Qualitative Results}
\label{sec:visualize}

\begin{figure}[h!]
\centering
\includegraphics[width=\linewidth]{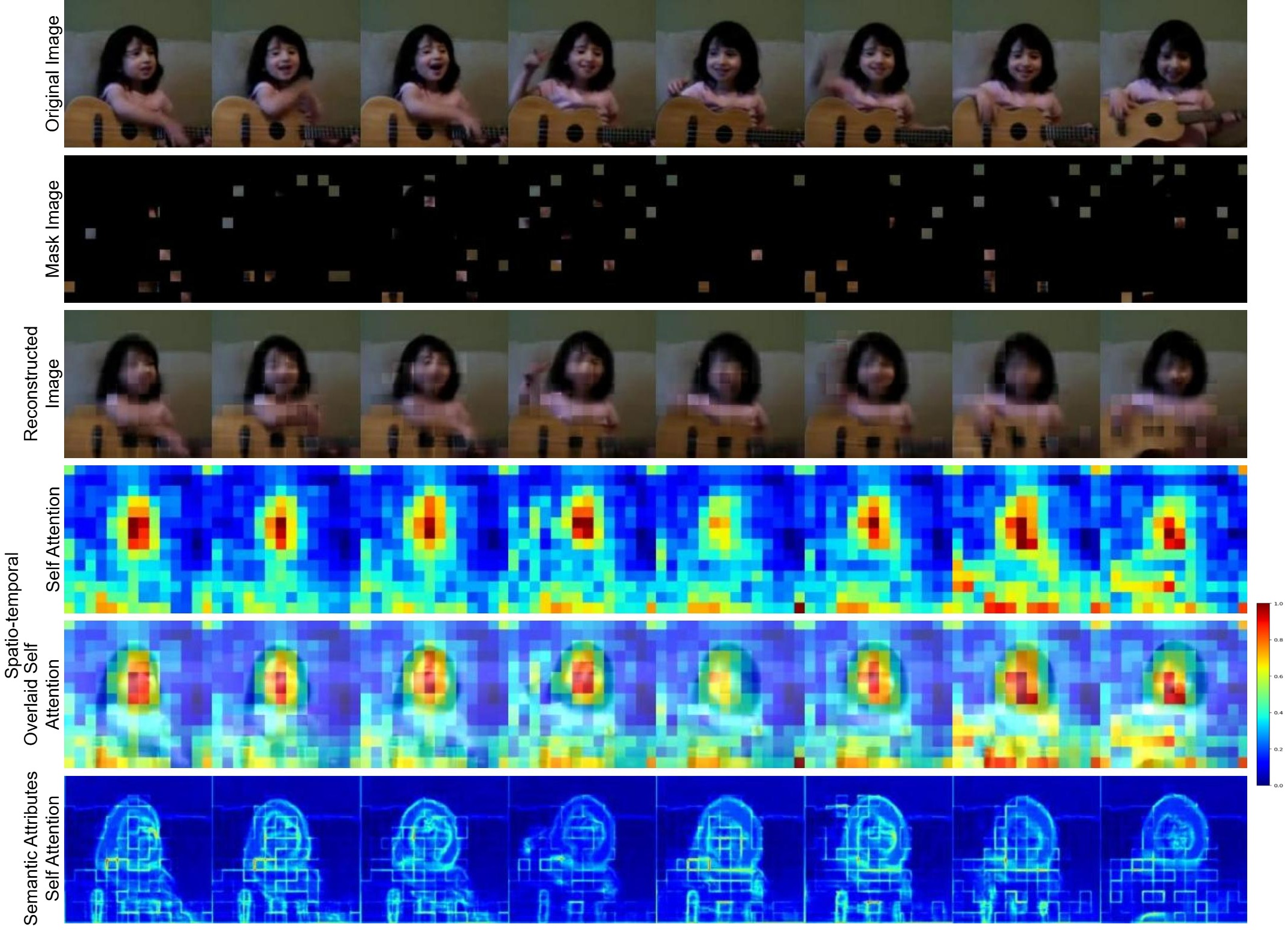}
\caption{An example self-attention maps visualization of our CrossVideoMAE on the K400 dataset.}
\label{fig:figure1}
\end{figure}

\begin{figure}[h!]
\centering
\includegraphics[width=\linewidth]{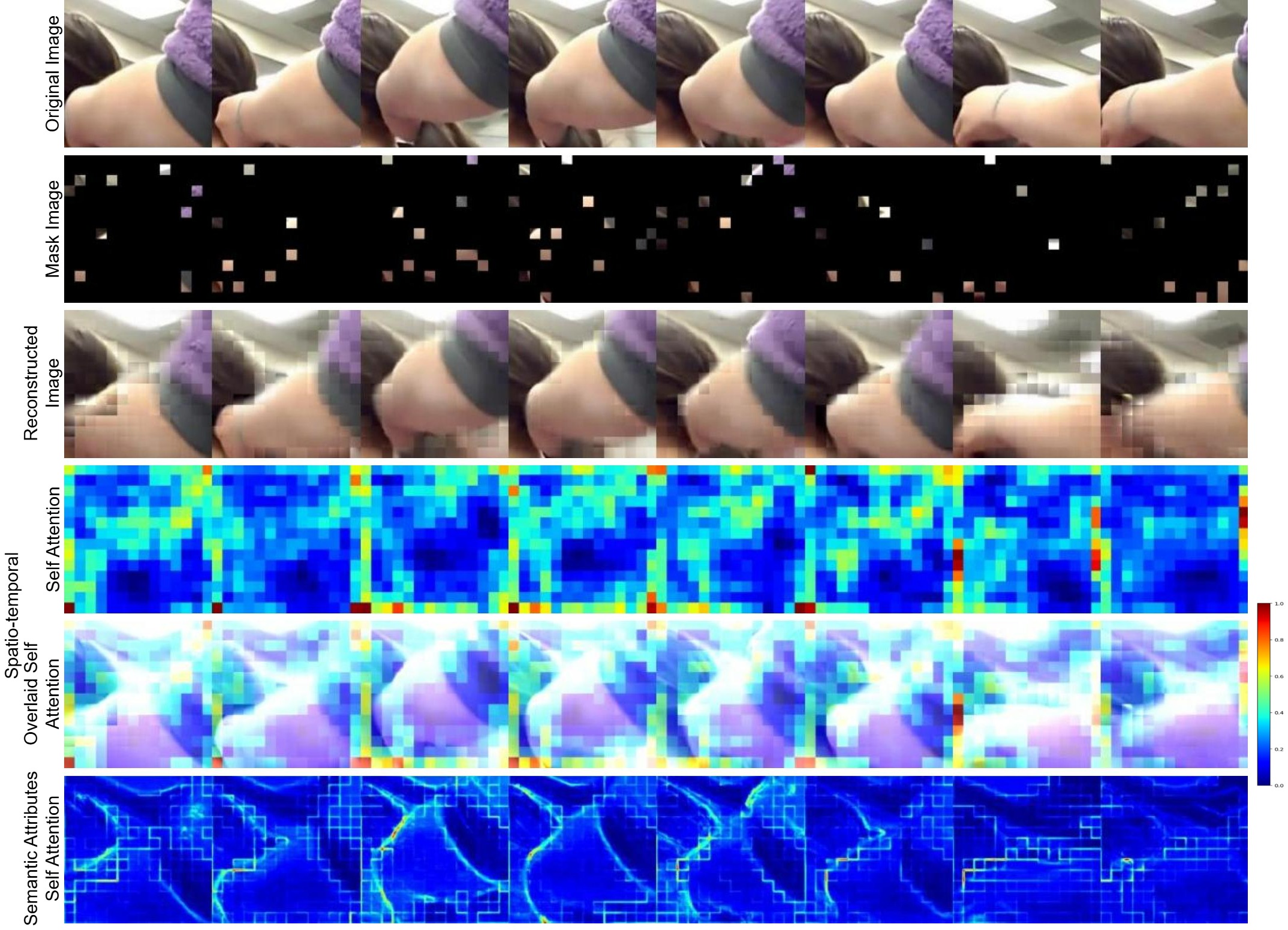}
\caption{An example self-attention maps visualization of our CrossVideoMAE on the K400 dataset for a masking ratio of 95\%.}
\label{fig:figure2}
\end{figure}

\begin{figure}[h!]
\centering
\includegraphics[width=\linewidth]{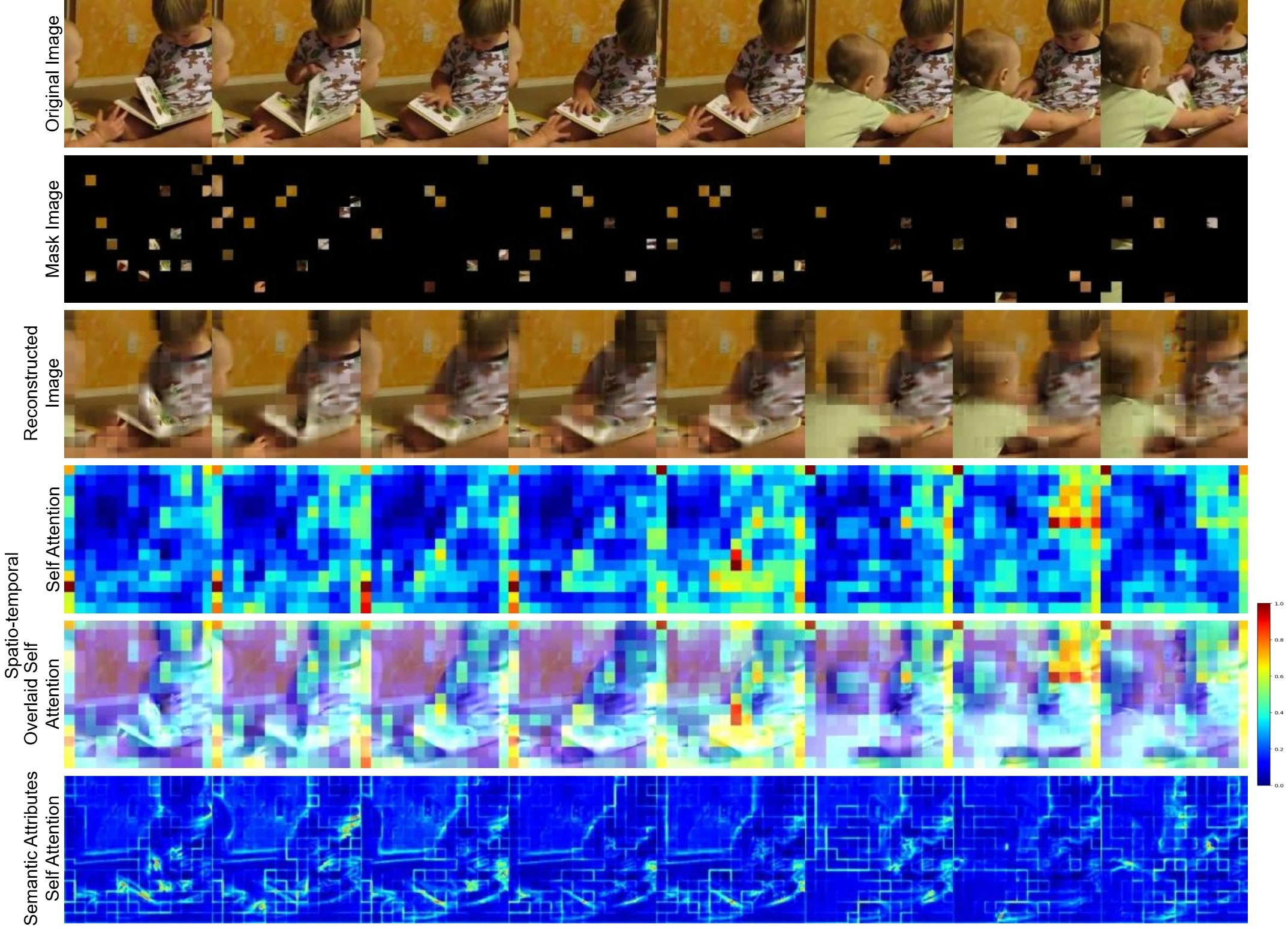}
\caption{An example self-attention maps visualization of our CrossVideoMAE on the K400 dataset for a masking ratio of 95\%.}
\label{fig:figure3}
\end{figure}

\begin{figure}[h!]
\centering
\includegraphics[width=\linewidth]{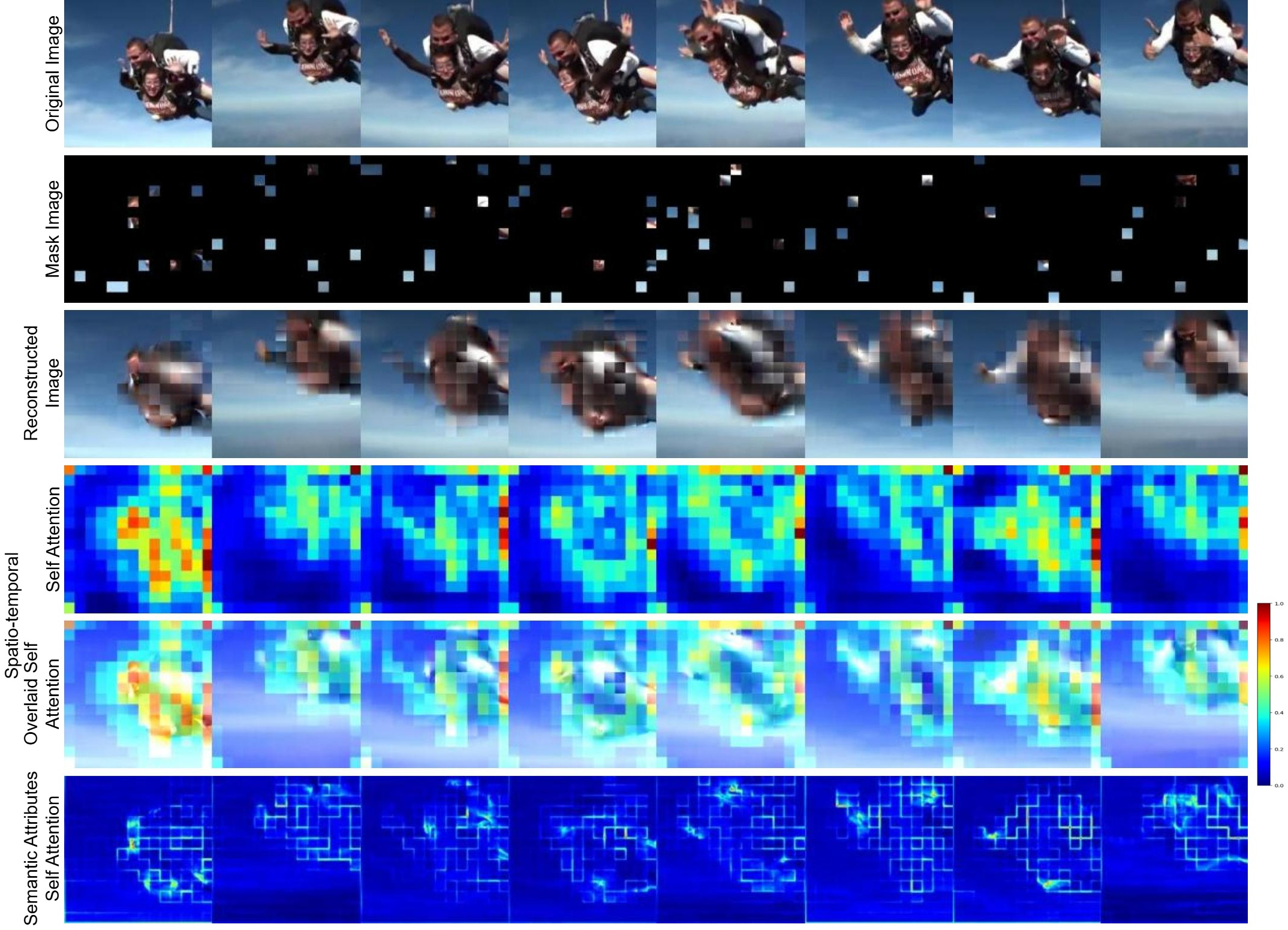}
\caption{An example self-attention maps visualization of our CrossVideoMAE on the K400 dataset for a masking ratio of 95\%.}
\label{fig:figure4}
\end{figure}

\begin{figure}[h!]
\centering
\includegraphics[width=\linewidth]{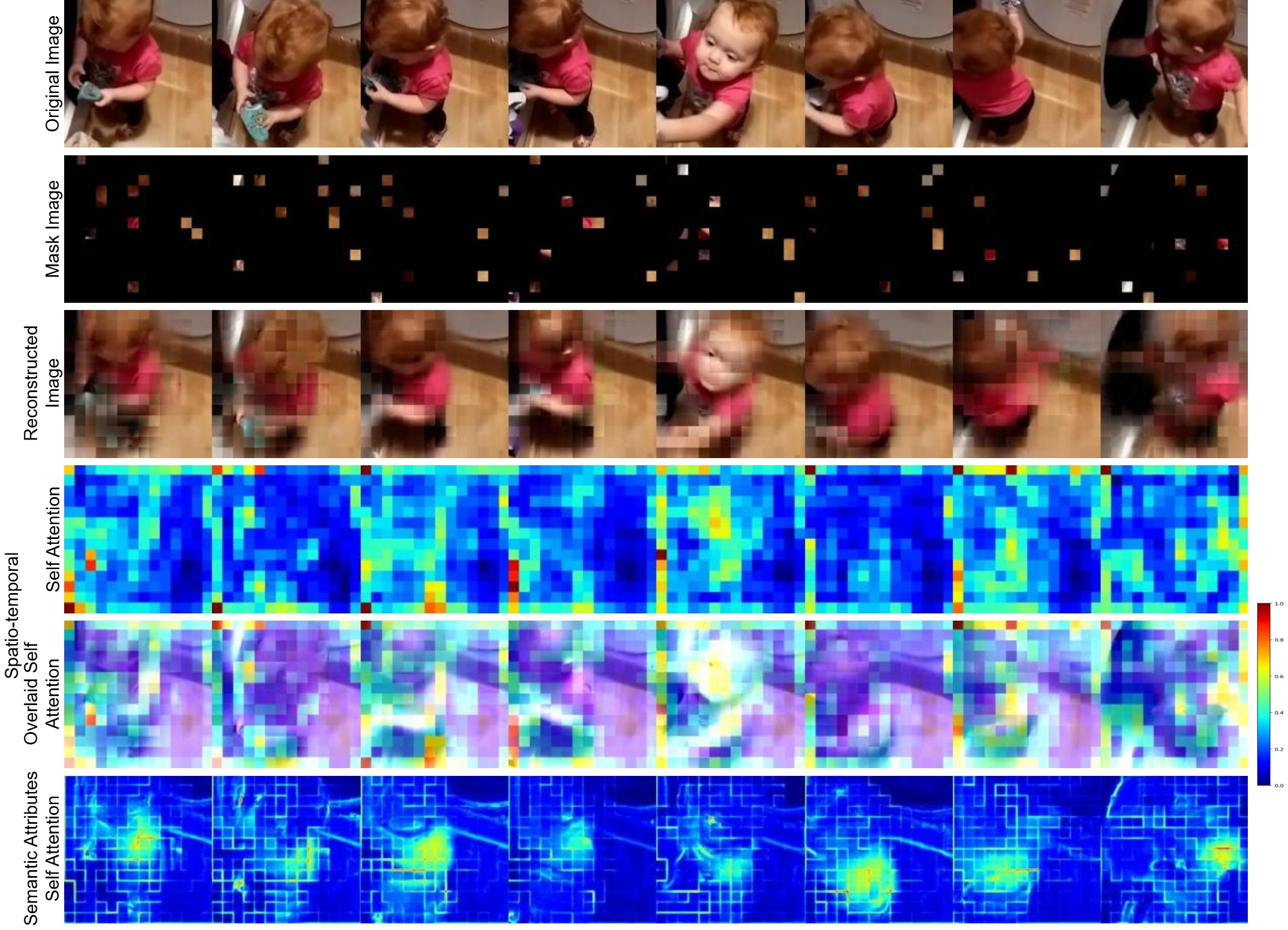}
\caption{An example self-attention maps visualization of our CrossVideoMAE on the K400 dataset for a masking ratio of 95\%.}
\label{fig:figure5}
\end{figure}

\begin{figure}[h!]
\centering
\includegraphics[width=\linewidth]{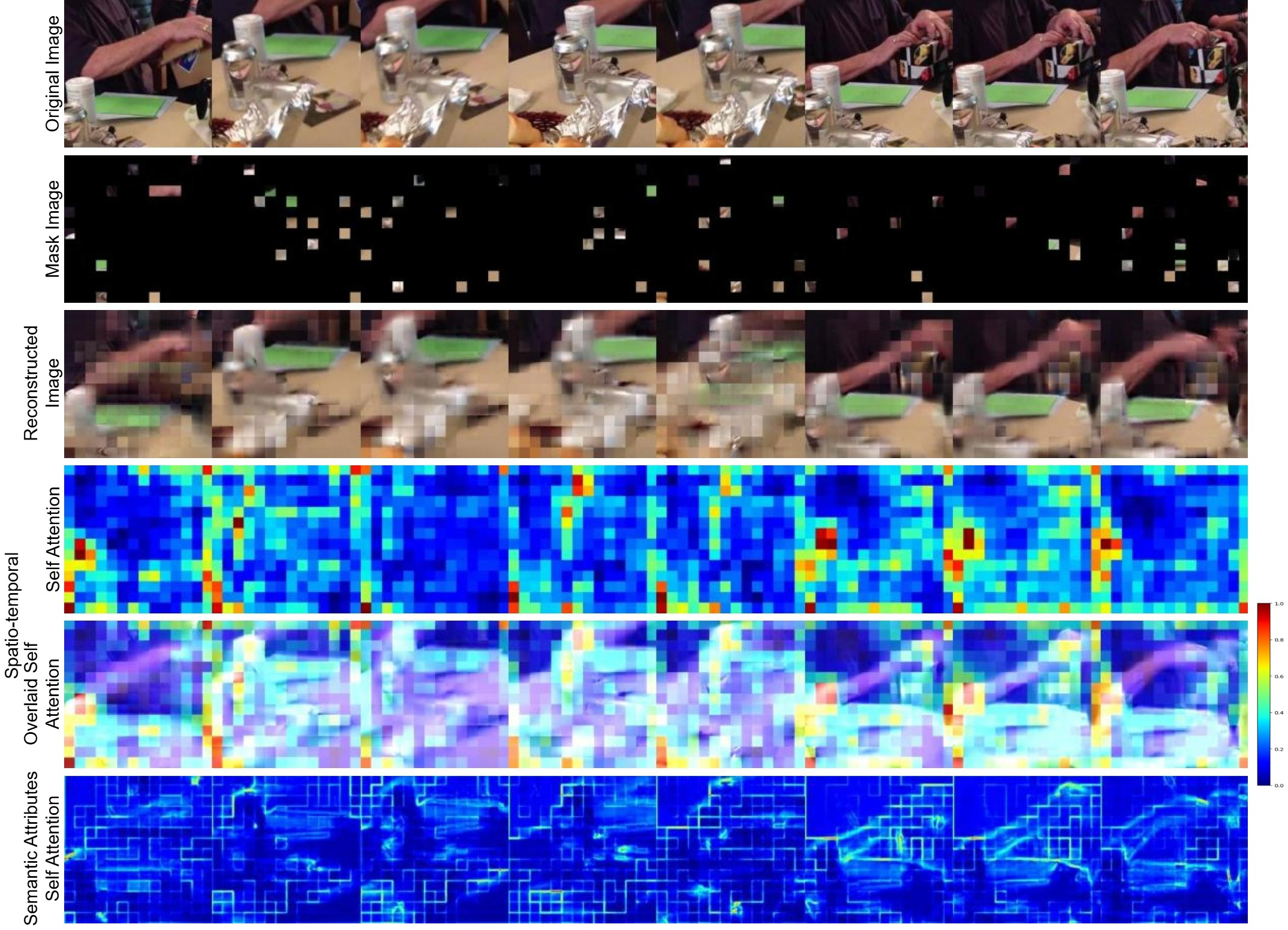}
\caption{An example self-attention maps visualization of our CrossVideoMAE on the K400 dataset for a masking ratio of 95\%.}
\label{fig:figure6}
\end{figure}

\begin{figure}[h!]
\centering
\includegraphics[width=\linewidth]{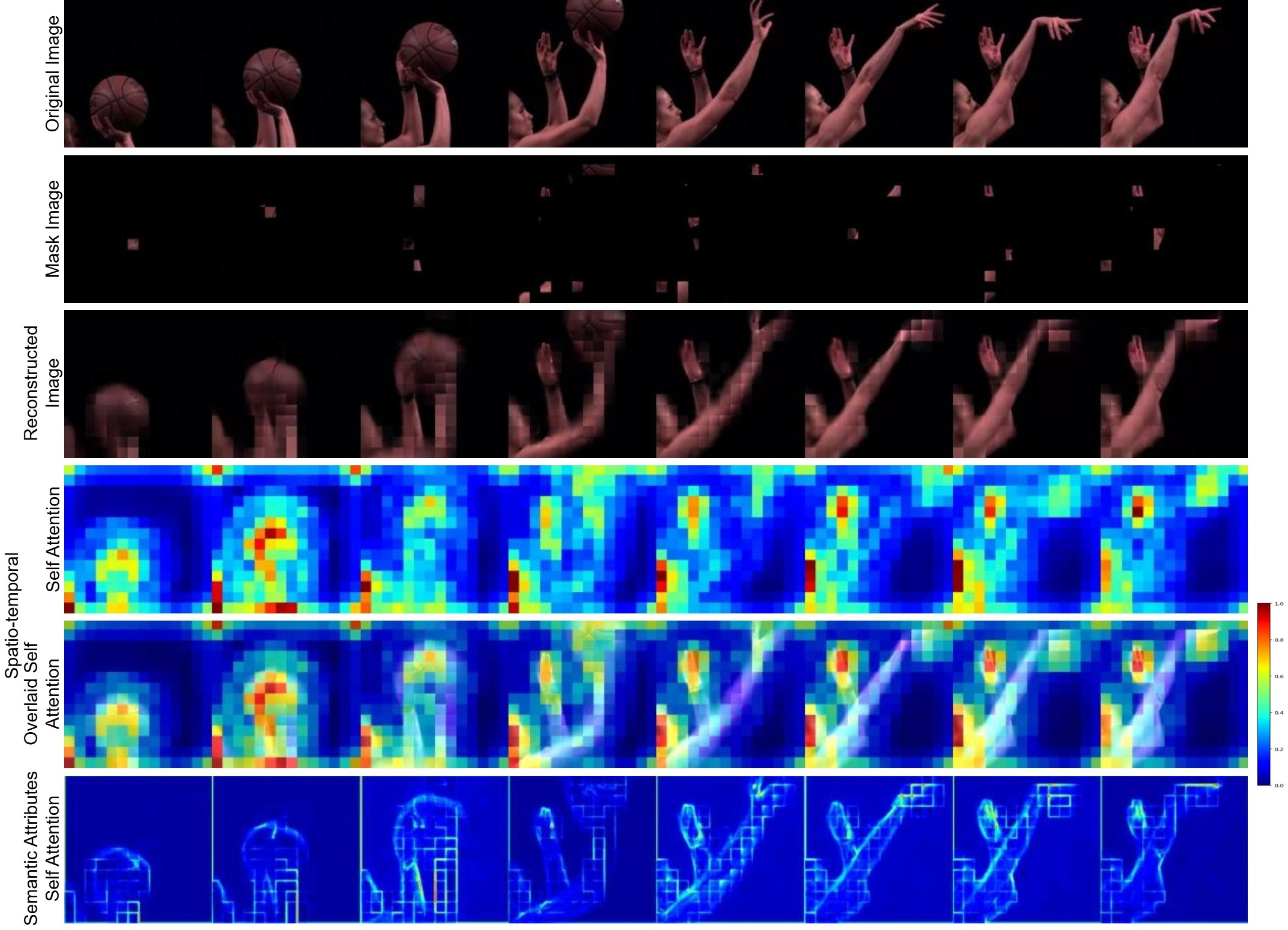}
\caption{An example self-attention maps visualization of our CrossVideoMAE on the K400 dataset for a masking ratio of 95\%.}
\label{fig:figure7}
\end{figure}

\begin{figure}[h!]
\centering
\includegraphics[width=\linewidth]{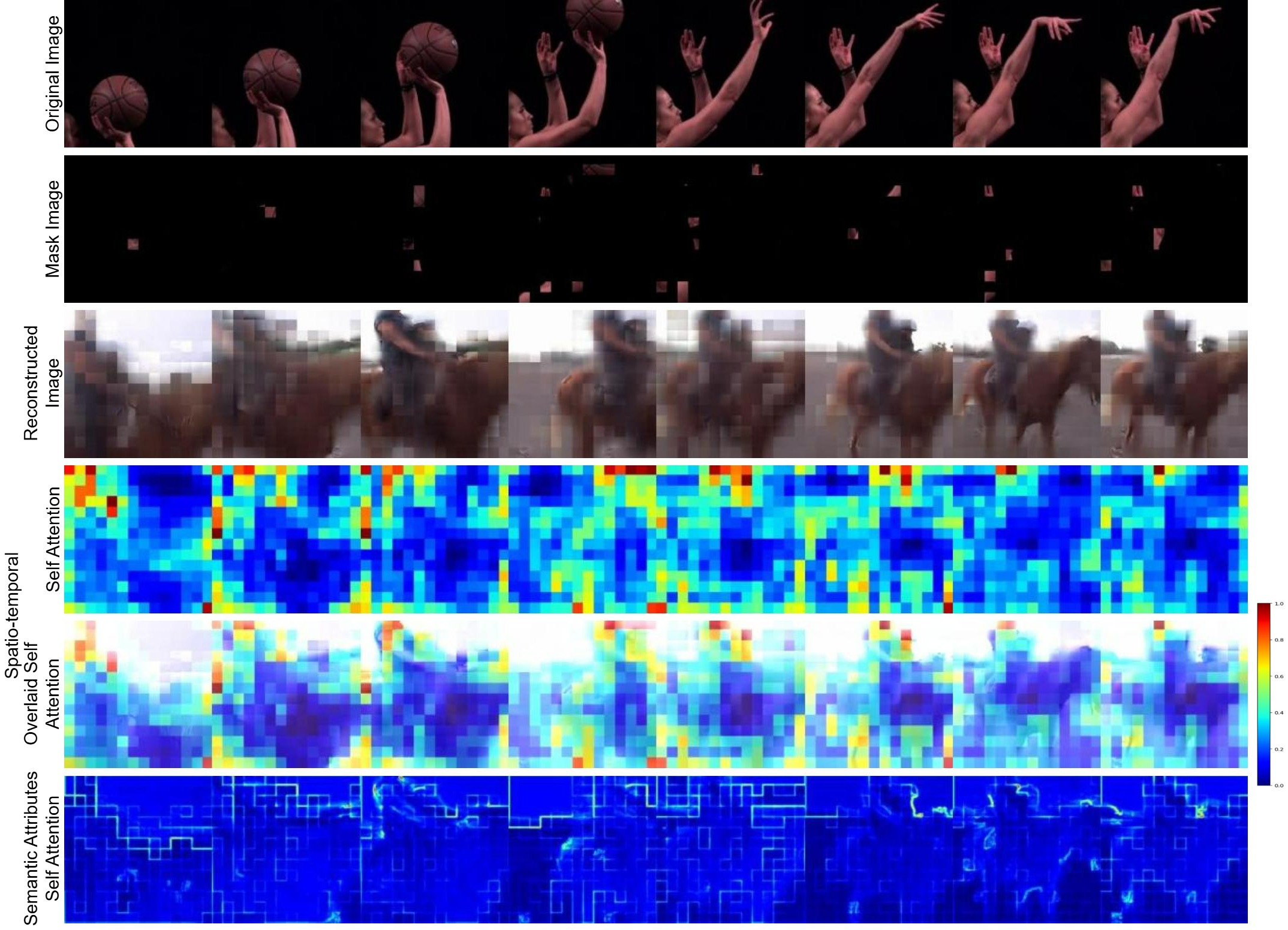}
\caption{An example self-attention maps visualization of our CrossVideoMAE on the K400 dataset for a masking ratio of 95\%.}
\label{fig:figure8}
\end{figure}

\begin{figure}[h!]
\centering
\includegraphics[width=\linewidth]{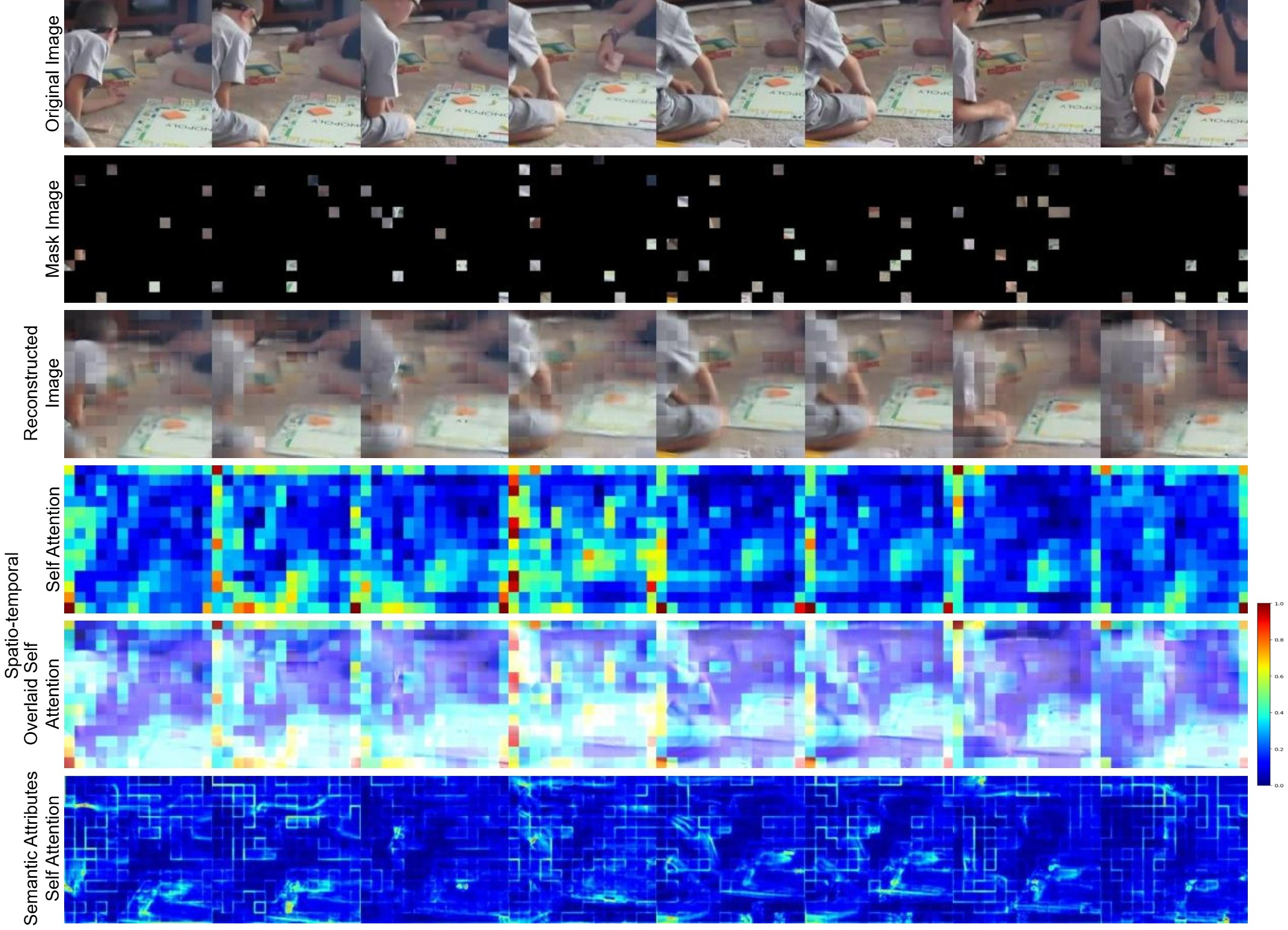}
\caption{An example self-attention maps visualization of our CrossVideoMAE on the K400 dataset for a masking ratio of 95\%.}
\label{fig:figure9}
\end{figure}

\begin{figure}[h!]
\centering
\includegraphics[width=\linewidth]{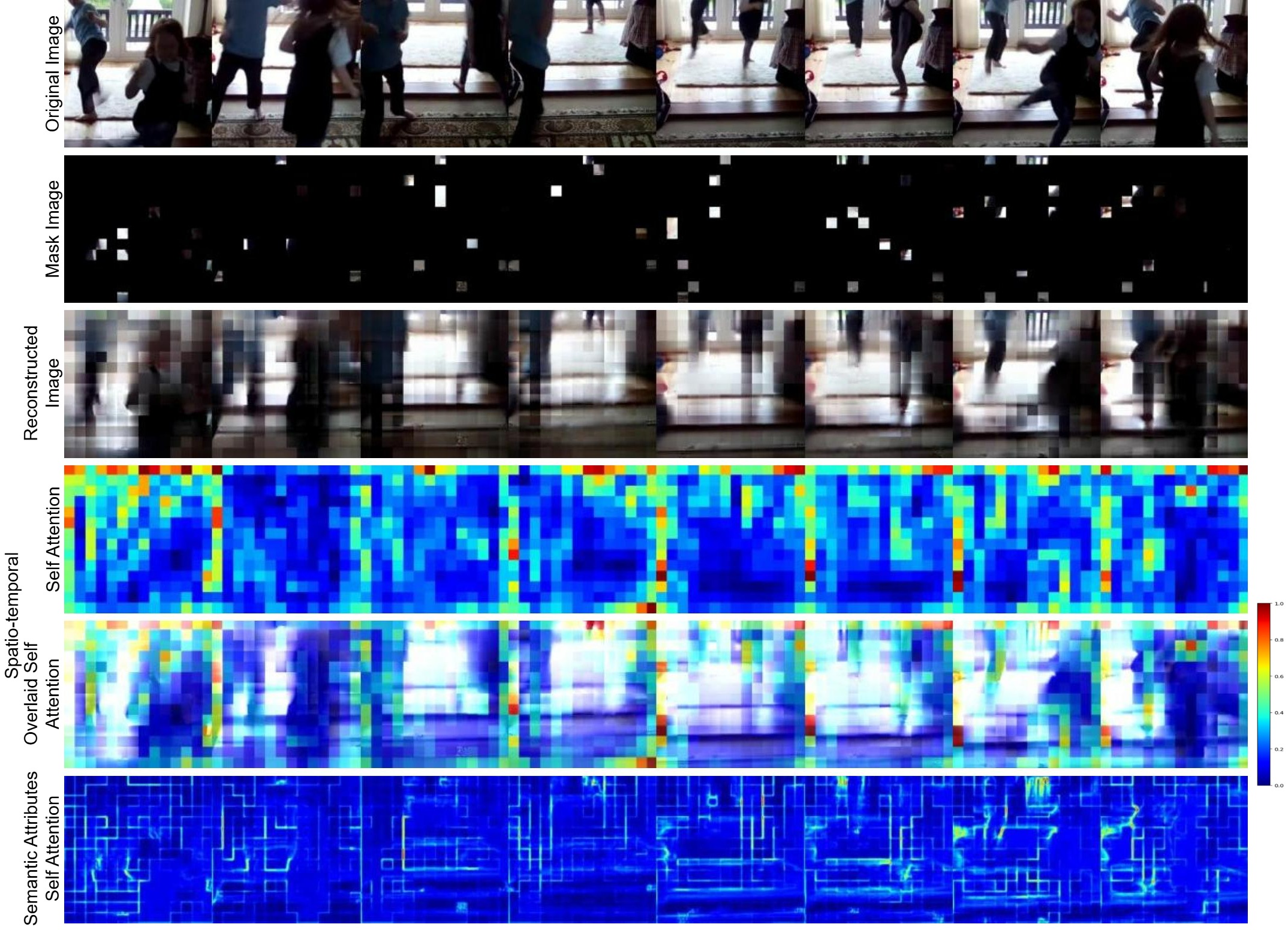}
\caption{An example self-attention maps visualization of our CrossVideoMAE on the K400 dataset for a masking ratio of 95\%.}
\label{fig:figure10}
\end{figure}

\begin{figure}[h!]
\centering
\includegraphics[width=\linewidth]{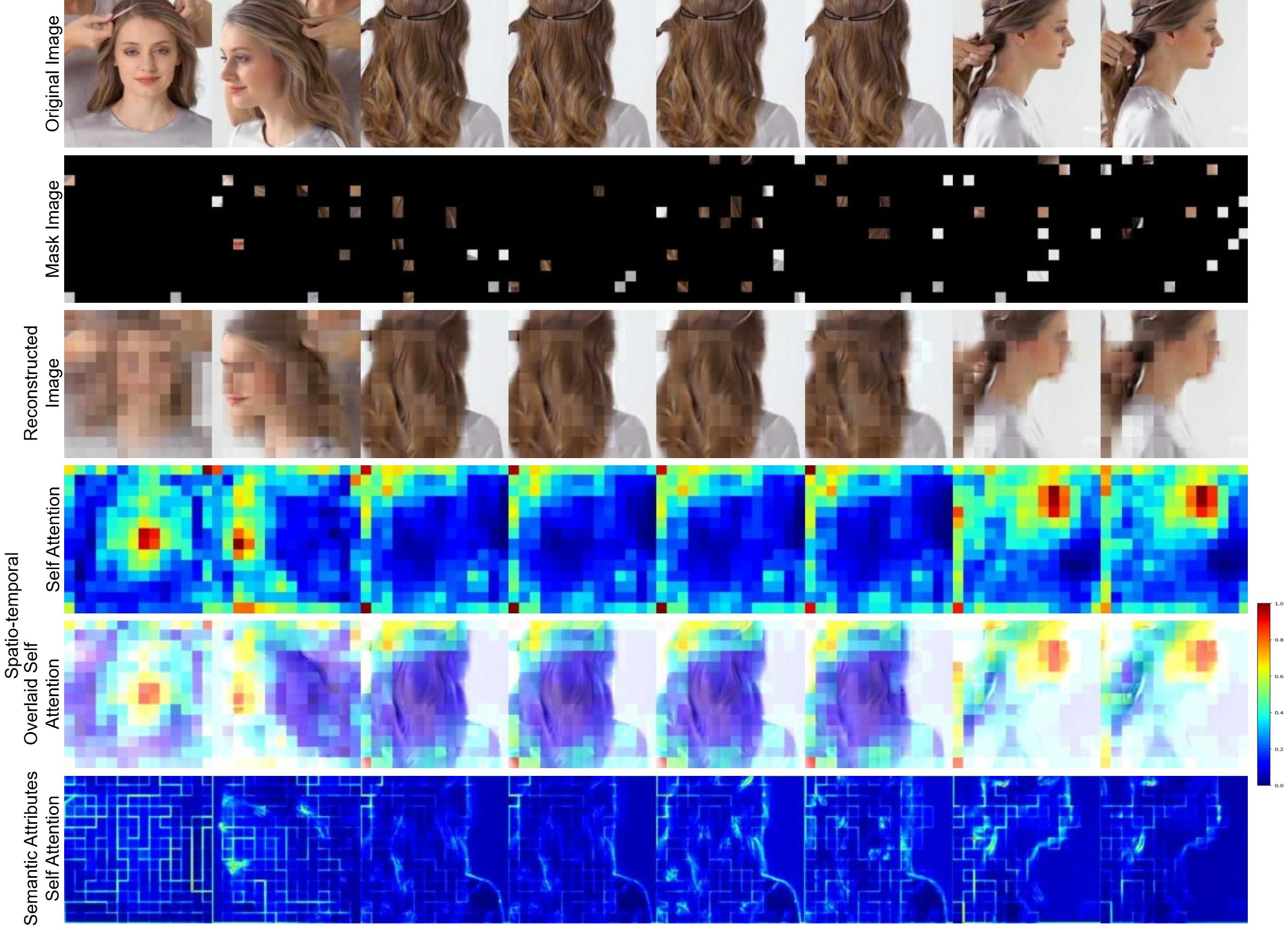}
\caption{An example self-attention maps visualization of our CrossVideoMAE on the K400 dataset for a masking ratio of 95\%.}
\label{fig:figure11}
\end{figure}

\begin{figure}[h!]
\centering
\includegraphics[width=\linewidth]{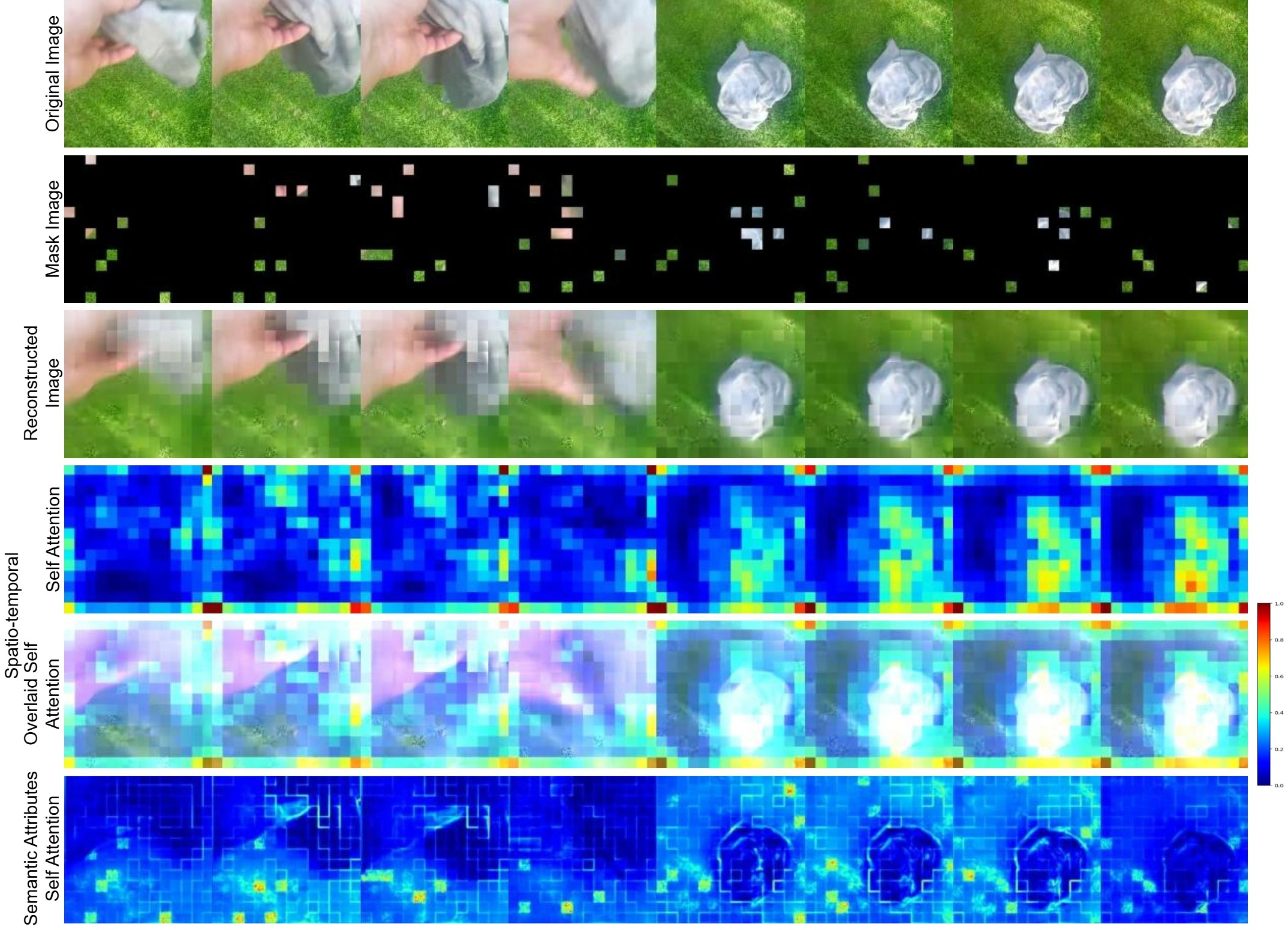}
\caption{An example self-attention maps visualization of our CrossVideoMAE on SSv2 dataset for a masking ratio of 95\%.}
\label{fig:figure12}
\end{figure}

\begin{figure}[h!]
\centering
\includegraphics[width=\linewidth]{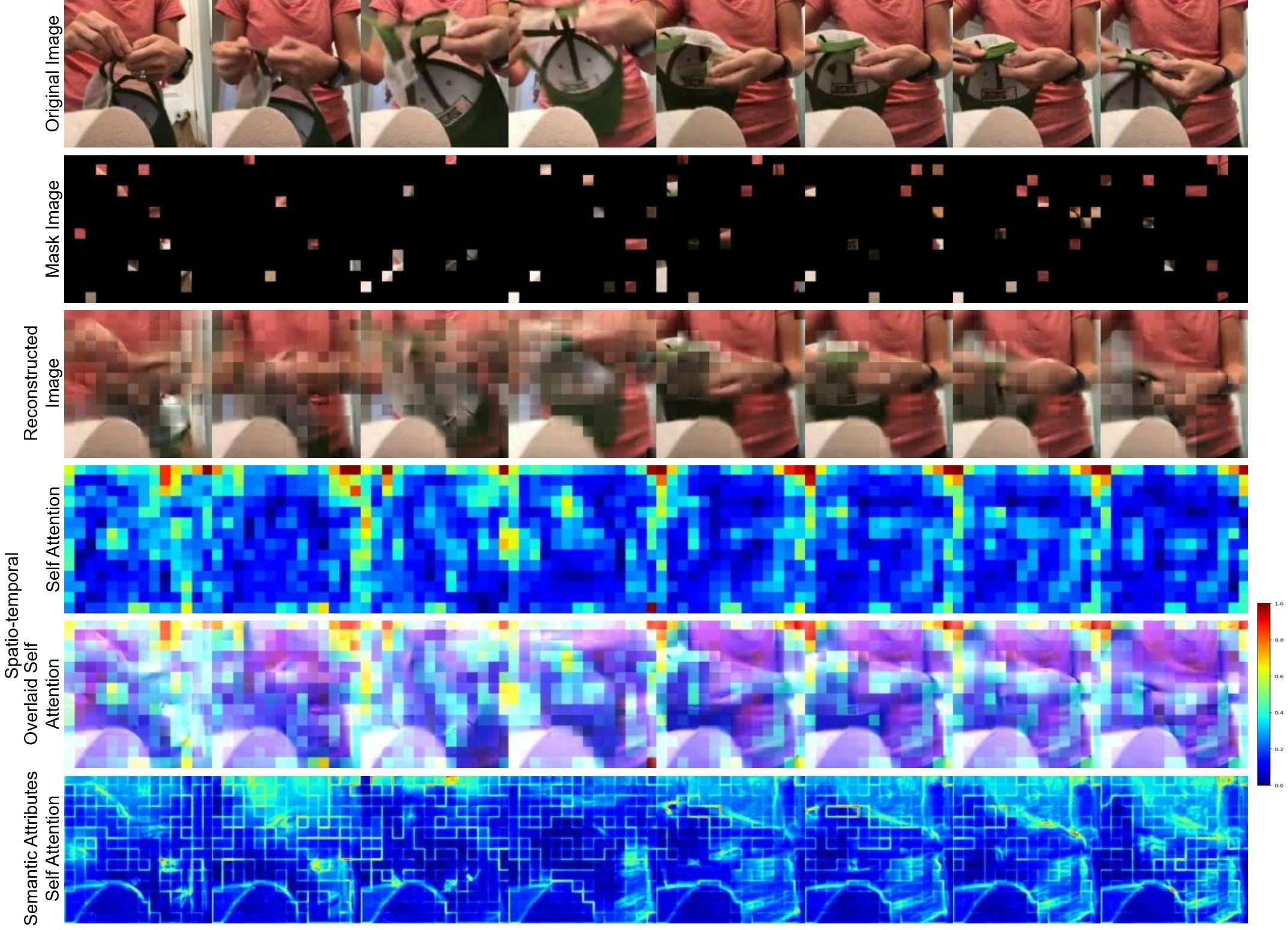}
\caption{An example self-attention maps visualization of our CrossVideoMAE on SSv2 dataset for a masking ratio of 95\%.}
\label{fig:figure13}
\end{figure}

\begin{figure}[h!]
\centering
\includegraphics[width=\linewidth]{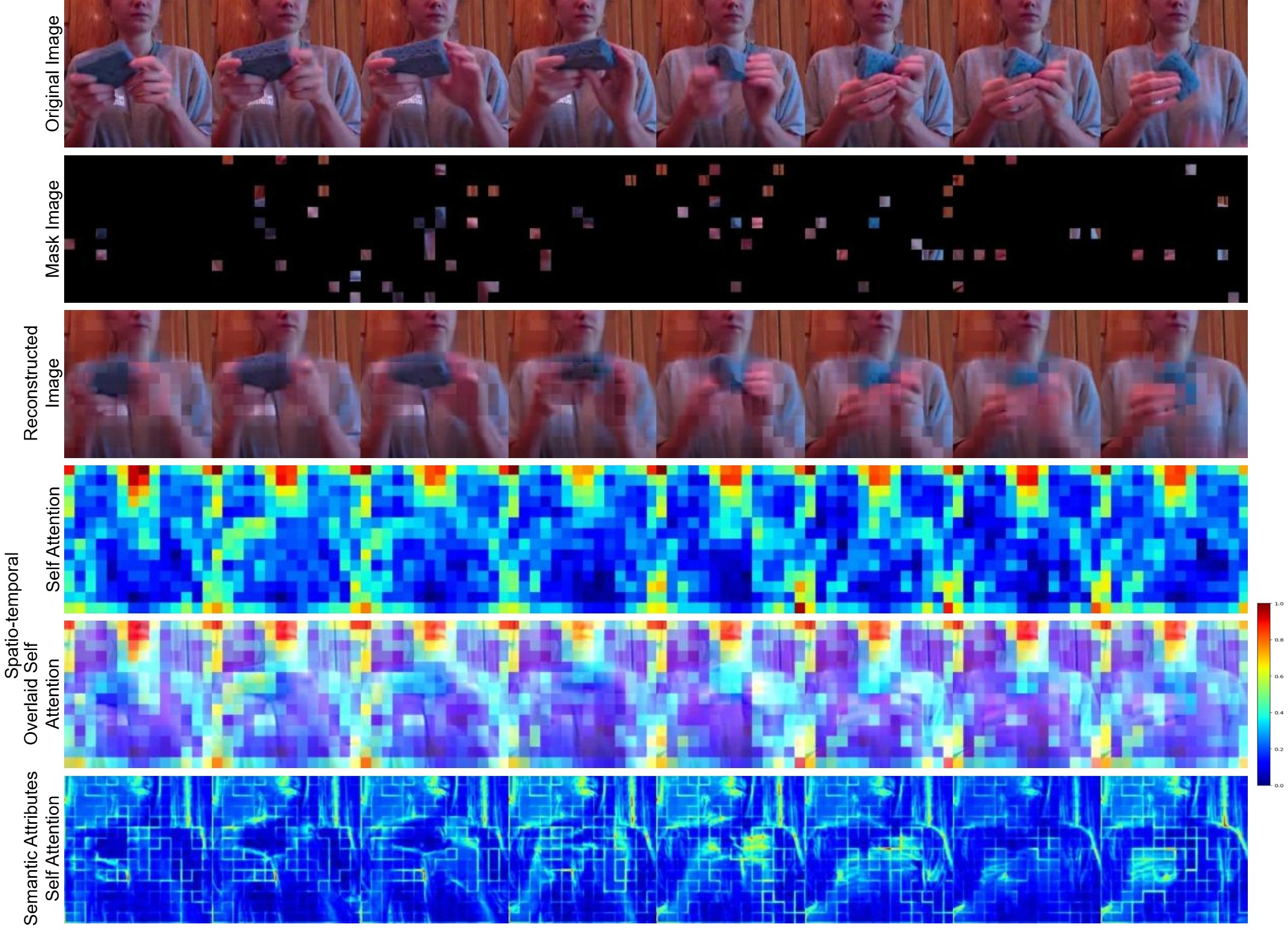}
\caption{An example self-attention maps visualization of our CrossVideoMAE on SSv2 dataset for a masking ratio of 95\%.}
\label{fig:figure14}
\end{figure}

\begin{figure}[h!]
\centering
\includegraphics[width=\linewidth]{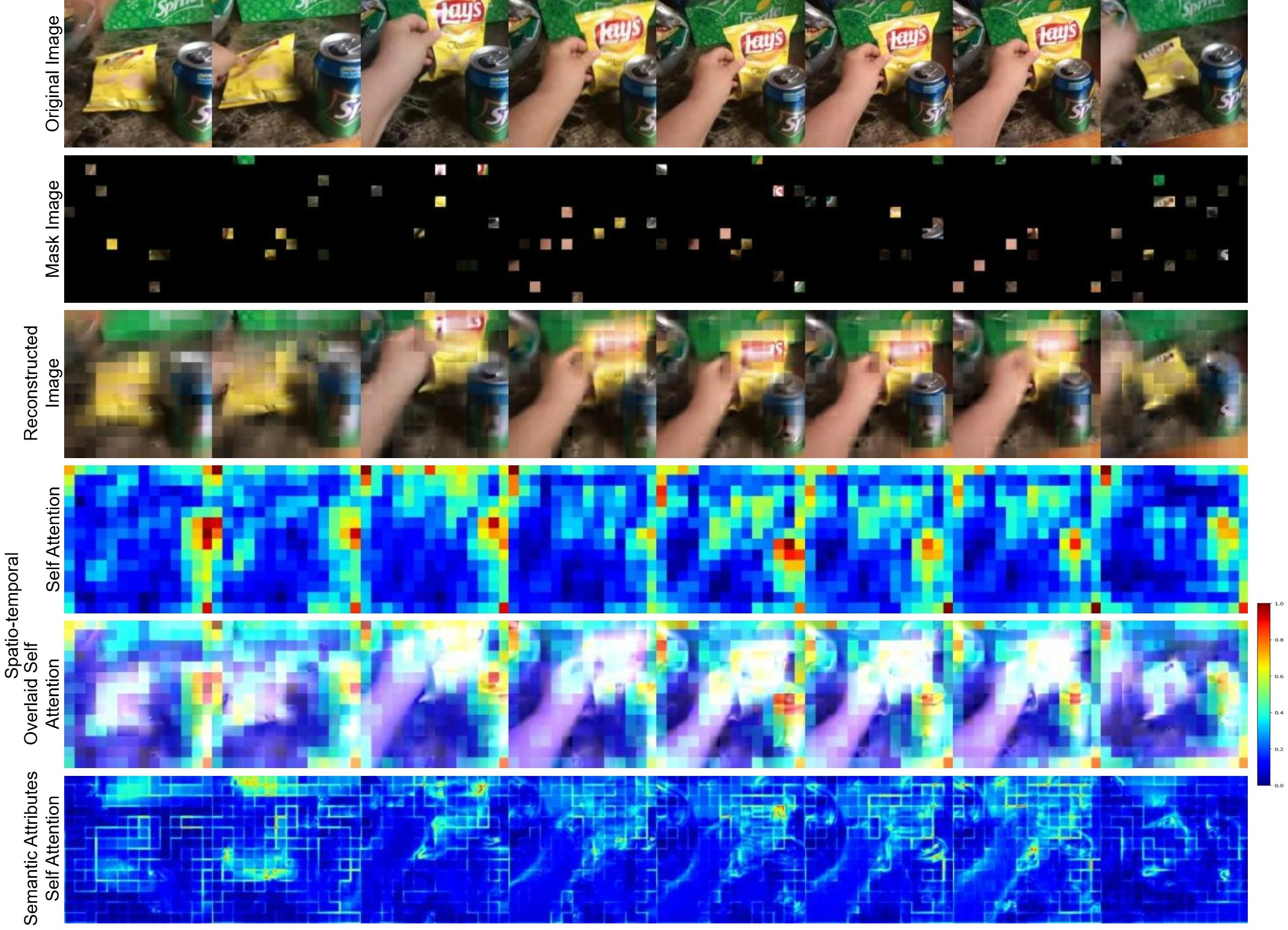}
\caption{An example self-attention maps visualization of our CrossVideoMAE on SSv2 dataset for a masking ratio of 95\%.}
\label{fig:figure15}
\end{figure}

\begin{figure}[h!]
\centering
\includegraphics[width=\linewidth]{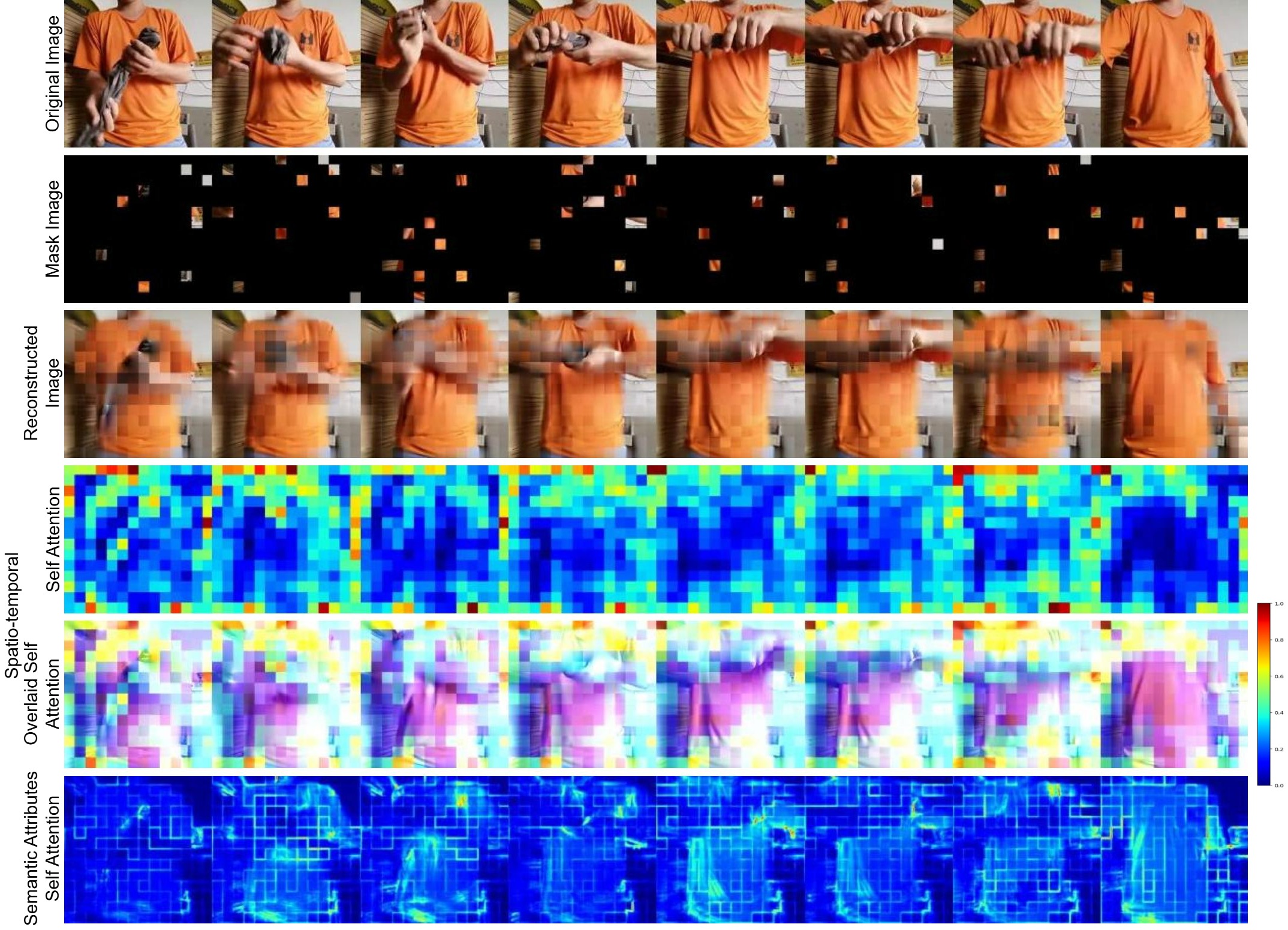}
\caption{An example self-attention maps visualization of our CrossVideoMAE on SSv2 dataset for a masking ratio of 95\%.}
\label{fig:figure16}
\end{figure}

\begin{figure}[t!]
\includegraphics[width=\linewidth]{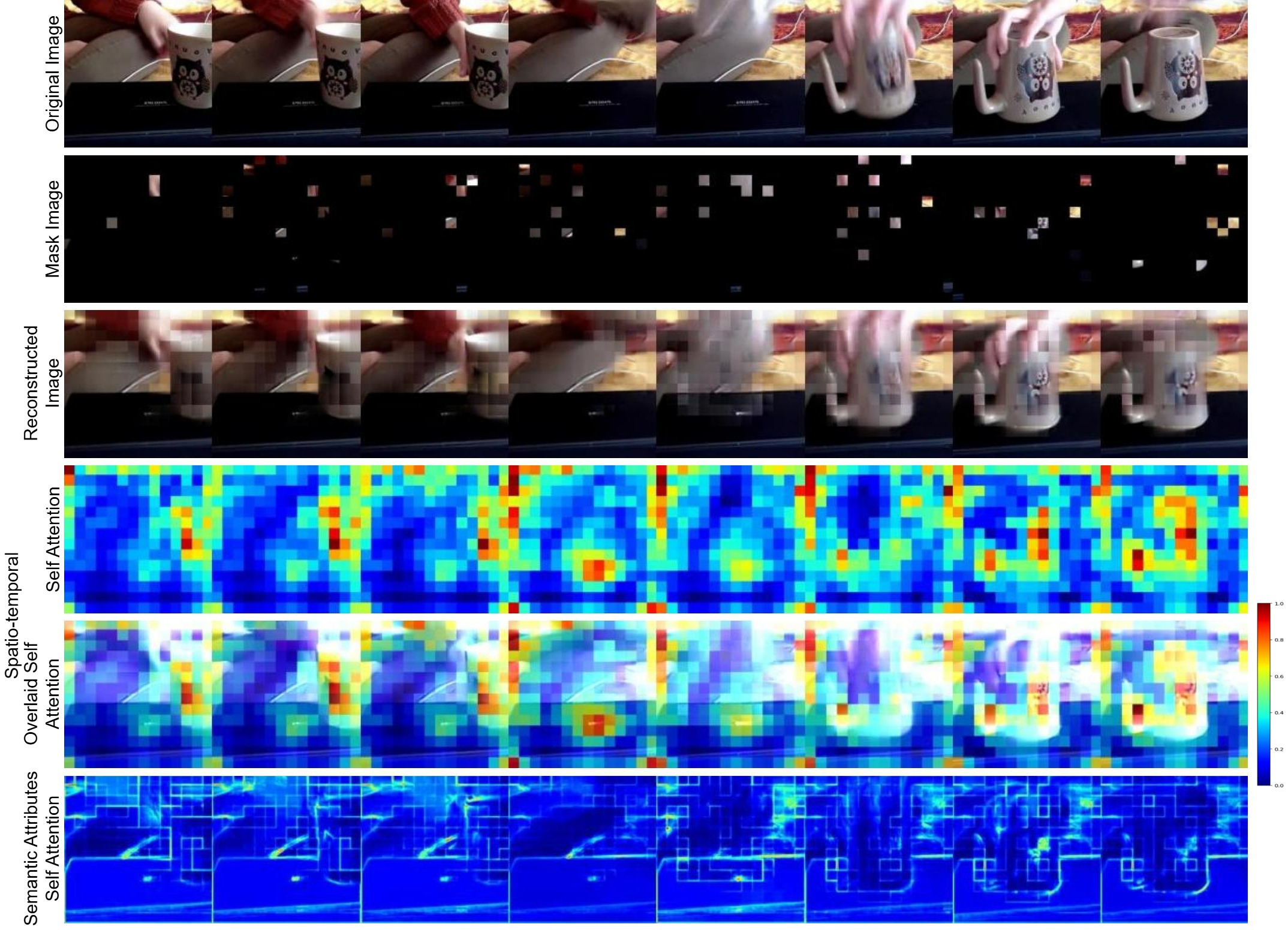}
\caption{An example self-attention maps visualization of our CrossVideoMAE on SSv2 dataset for a masking ratio of 95\%.}
\label{fig:figure17}

\centering
\includegraphics[width=\linewidth]{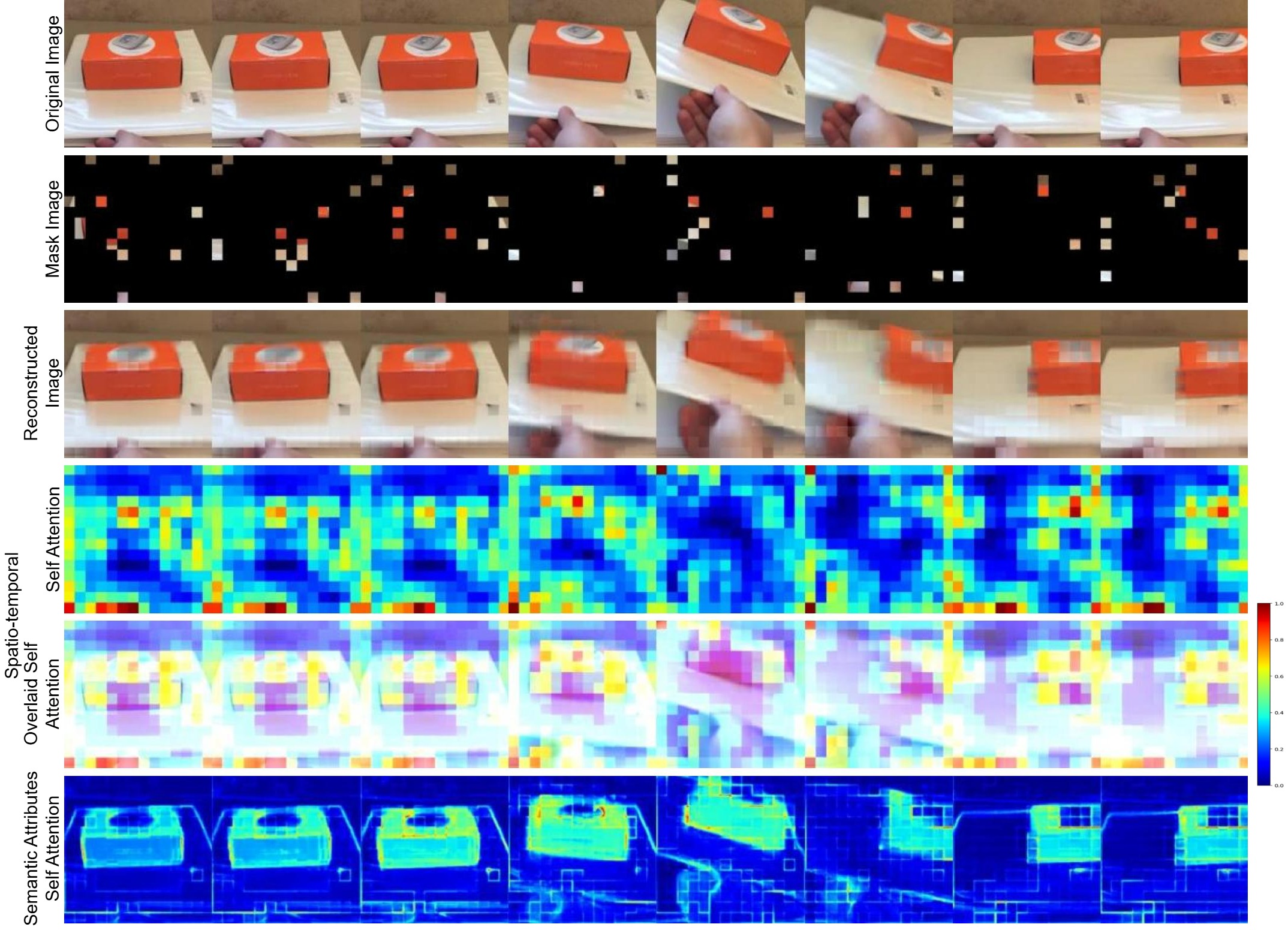}
\caption{An example self-attention maps visualization of our CrossVideoMAE on SSv2 dataset for a masking ratio of 95\%.}
\label{fig:figure18}

\includegraphics[width=\linewidth]{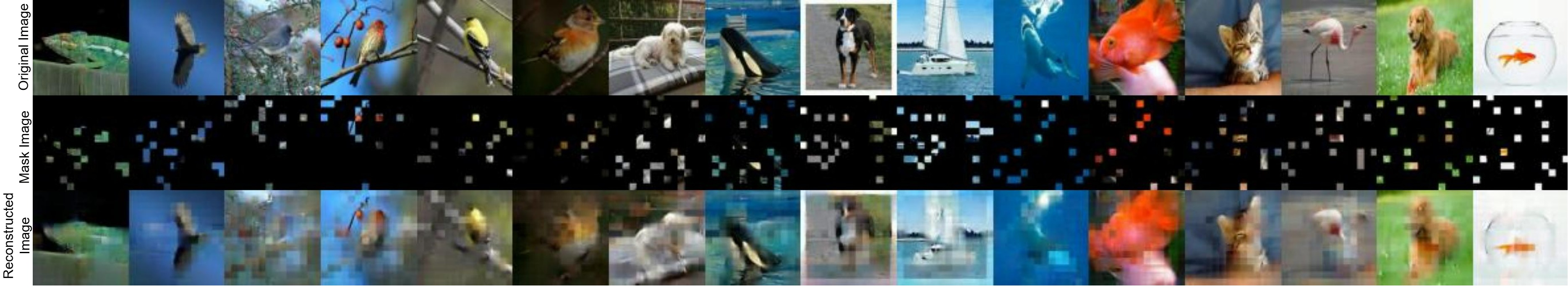}
\caption{\textbf{Additional reconstruction visualizations.} using CrossVideoMAE on the IN-1K image dataset. We show the model predictions for a masking ratio of 90\%.}
\label{fig:figure19}
\end{figure}

To further understand how the proposed CrossVideoMAE approach effectively captures rich spatiotemporal representations and semantic attributes in videos, we analyze the self-attention maps for reconstructed samples from randomly selected additional videos in the K400(Fig.~\ref{fig:figure1}–\ref{fig:figure11}) and the SSv2 (Fig.~\ref{fig:figure12}–\ref{fig:figure18}) validation set and additional images in the IN-1K (Fig.~\ref{fig:figure19}) validation set. Even under high masking ratios, CrossVideoMAE demonstrates the ability to produce satisfying reconstruction results. These examples highlight the capability of CrossVideoMAE to learn and preserve complex spatiotemporal structures and semantic attributes in video data, underscoring its robustness and effectiveness in representation learning. 

For instance, in Fig.~\ref{fig:figure1}, the spatiotemporal representations are primarily concentrated in the central and lower regions of each frame, specifically focusing on the girl’s hand and lip movements while playing the guitar. Accurately reconstructing these regions is challenging, as evident in the third and sixth rows. The proposed CrossVideoMAE leverage difference between masking ratios applied to both branches across sampled frames and videos to effectively learn representations. This process allows the model to utilize visible tokens from both the sampled frame and the broader video context. Similar observations can be made for the other examples, further validating the capability of CrossVideoMAE to capture nuanced spatiotemporal and semantic details in video data.

Similar observations can be made for the other examples, further reinforcing the effectiveness of CrossVideoMAE in capturing nuanced spatiotemporal and semantic representations across diverse video samples. Upon acceptance, we plan to release additional \textbf{GIF} visualizations, alongside the code, on GitHub to provide a more comprehensive understanding of the proposed method's capabilities.

\noindent \textit{These results are for the default setting pre-training.}

\end{document}